\documentclass[sn-mathphys,Numbered]{sn-jnl}% Math and Physical Sciences Reference Style
\usepackage{graphicx}%
\usepackage{multirow}
\usepackage{amsmath,amssymb,amsfonts}%
\usepackage{amsthm}%
\usepackage{mathrsfs}%
\usepackage[title]{appendix}%
\usepackage{xcolor}%
\usepackage{textcomp}%
\usepackage{manyfoot}%
\usepackage{booktabs}\let\cline\cmidrule
\usepackage{algorithm}%
\usepackage{algorithmicx}%
\usepackage{algpseudocode}%
\usepackage{listings}%
\usepackage[edges]{forest} % for tree structures
\usepackage{amsmath}
\usepackage{adjustbox}
\usepackage{makecell}
\usepackage[inline]{enumitem}
\usepackage{rotating}

\begin{document}

\title[Article Title]{Beyond Efficiency: A Systematic Survey of Resource-Efficient Large Language Models}


\author[1]{\fnm{Guangji} \sur{Bai}}\email{guangji.bai@emory.edu}

\author[2]{\fnm{Zheng} \sur{Chai}}\email{dub6yh@virginia.edu}

\author[1]{\fnm{Chen} \sur{Ling}}\email{chen.ling@emory.edu}

\author[1]{\fnm{Shiyu} \sur{Wang}}\email{shiyu.wang@emory.edu}

\author[1]{\fnm{Jiaying} \sur{Lu}}\email{jiaying.lu@emory.edu}

\author[3]{\fnm{Nan} \sur{Zhang}}\email{njz5124@psu.edu}

\author[1]{\fnm{Tingwei} \sur{Shi}}\email{tshi30@emory.edu}

\author[1]{\fnm{Ziyang} \sur{Yu}}\email{zyu31@emory.edu}

\author[1]{\fnm{Mengdan} \sur{Zhu}}\email{mengdan.zhu@emory.edu}

\author[1]{\fnm{Yifei} \sur{Zhang}}\email{yifei.zhang2@emory.edu}

\author[1]{\fnm{Xinyuan} \sur{Song}}\email{xsong30@emory.edu}

\author[1]{\fnm{Carl} \sur{Yang}}\email{j.carlyang@emory.edu}

\author[2]{\fnm{Yue} \sur{Cheng}}\email{mrz7dp@virginia.edu}

\author*[1]{\fnm{Liang} \sur{Zhao}}\email{liang.zhao@emory.edu}

\affil*[1]{\orgdiv{Department of Computer Science}, \orgname{Emory University}, \orgaddress{\street{201 Dowman Dr}, \city{Atlanta}, \postcode{30322}, \state{GA}, \country{United States}}}

\affil[2]{\orgdiv{School of Data Science and Department of Computer Science}, \orgname{University of Virginia}, \orgaddress{\street{1827 University Avenue}, \city{Charlottesville}, \postcode{22904}, \state{VA}, \country{United States}}}

\affil[3]{\orgdiv{College of Information Sciences and Technology}, \orgname{Pennsylvania State University}, \orgaddress{\street{201 Old Main}, \city{University Park}, \postcode{16802}, \state{PA}, \country{United States}}}


\abstract{The burgeoning field of Large Language Models (LLMs), exemplified by sophisticated models like OpenAI's ChatGPT, represents a significant advancement in artificial intelligence. These models, however, bring forth substantial challenges in high consumption of
%terms of 
computational, memory, energy, and financial resources, especially in environments with limited resource capabilities. This survey aims to systematically address these challenges by reviewing a broad spectrum of techniques designed to enhance the resource efficiency of LLMs. We categorize methods based on their optimization focus—covering computational, memory, energy, financial, and network 
%communication 
resources—and their applicability across various stages of an LLM's lifecycle, including architecture design, pre-training, fine-tuning, and system design. Additionally, the survey introduces a nuanced categorization of resource efficiency techniques by their specific resource types, which uncovers the intricate relationships and mappings between various resources and corresponding optimization techniques. A standardized set of evaluation metrics and datasets is also presented to facilitate consistent and fair comparisons across different models and techniques. By offering a comprehensive overview of the current state-of-the-art and identifying open research avenues, this survey serves as a foundational reference for researchers and practitioners, aiding them in developing more sustainable and efficient LLMs in a rapidly evolving landscape. To make the field more accessible to interested beginners, we not only make a systematic review of existing works and a highly structured taxonomy of resource-efficient LLMs but also release a website including a constantly-updated paper list~\url{https://github.com/tiingweii-shii/Awesome-Resource-Efficient-LLM-Papers}.}




\maketitle

\section{Introduction}
\label{sec: introduction}

In recent years, Large Language Models (LLMs)~\cite{openai2023gpt,touvron2023llama} have achieved significant advancements, redefining the frontier of artificial intelligence. These models, such as OpenAI's GPT-3 with its impressive 175 billion parameters, represent a quantum leap in complexity and capability~\cite{floridi2020gpt}. The trend in LLM development is toward ever-increasing model sizes, with some recent entrants boasting upwards of hundreds of billions of parameters~\cite{zhang2022opt,scao2022bloom,chowdhery2023palm}. This scale amplifies their utility across a spectrum of applications, from intelligent 
chatbots to intricate data analyses, and even facilitating research in diverse domains. However, the exponential growth in model sizes presents a huge demand for various resources
%presents profound costs in various resources 
(e.g., computation, energy, memory)~\cite{chien2023reducing,samsi2023words,vsakota2023fly}. The immense resource requirements to train or deploy such extensive models can be cost-prohibitive, particularly in resource-constrained environments like academic labs or the medical sector, which do not have access to the vast computational resources of major IT conglomerates. Additionally, the environmental impact is a growing concern, as the extensive GPU usage for training these models translates to significant electricity consumption and increasing carbon dioxide emissions~\cite{chien2023reducing}. Addressing these challenges requires a focused effort on enhancing the resource efficiency of LLMs at every stage of their lifecycle.

Defining \emph{Resource-Efficient LLMs} requires an understanding of the critical resources involved in the lifecycle of LLMs. In this survey, we systematically categorize the essential resources into five key categories: \emph{computation, memory, energy, money, and communication cost}. Computation refers to the processing power necessary to train and run these models; memory encompasses the data storage capacity required; energy denotes the electricity consumed during operation; financial resources pertain to the monetary investment needed for infrastructure and ongoing costs; and communication cost involves the bandwidth and latency in data transfer during training and inference. Efficiency in this context is characterized by the ratio of these resources invested to the output produced, with a more efficient system being one that yields the same level of output while consuming fewer resources. A resource-efficient LLM, therefore, is designed to maximize performance and capabilities while minimizing resource expenditure across all these dimensions, thereby enabling more sustainable and accessible AI solutions.

Resource efficiency in LLMs is a crucial and complex area that demands innovative solutions to address significant challenges. These challenges, more pronounced in LLMs than in smaller neural networks like CNNs and MLPs, arise from the unique scale and complexity of LLMs. We outline these challenges from various key perspectives:

\begin{itemize}[leftmargin=*]

    \item \textbf{[Model]} 1. Low parallelism in auto-regressive generation: Auto-regressive token generation, the predominant method in LLMs, suffers from significant latency due to poor parallelism~\cite{kaddour2023challenges}. This is especially problematic for large model sizes or extensive input lengths, hindering efficient processing in both training and inference.
    2. Quadratic complexity in self-attention layers: The multi-head self-attention layer in LLMs exhibits quadratic complexity with respect to the input sequence length~\cite{vaswani2017attention}. This complexity creates a computational bottleneck as the input length increases, limiting the practical input sequence length LLMs can handle efficiently.

    \item \textbf{[Theory]} 1. Scaling laws and diminishing returns: theoretical insights into scaling laws for neural networks, particularly LLMs, suggest that as models become larger, the benefits in performance improvement per parameter added diminish~\cite{kaplan2020scaling}. This phenomenon raises questions about the optimal size of LLMs and the balance between resource investment and performance gain. 2. Generalization and overfitting: Theoretical work on generalization in machine learning is particularly relevant for LLMs~\cite{ying2019overview,tirumala2022memorization}. Understanding the limits of what large models can generalize from training data and the risks of overfitting is crucial for developing more resource-efficient models.

    \item \textbf{[System]} Given the substantial model size of LLMs and the vast training datasets, fitting them into the memory of a single GPU/TPU is unfeasible~\cite{rajbhandari2020zero,shoeybi2019megatron}. Consequently, intricate system designs become crucial to optimize the training process for LLMs and successfully accomplish the task. Furthermore, the system design gains increased significance due to the latency and throughput requirements associated with the inference tasks of LLMs, particularly when taking into account user experience and the constraints of a limited cost budget~\cite{aminabadi2022deepspeed, sheng2023flexgen}.

    \item \textbf{[Ethics]} 1. Dependence on large and proprietary training data: Many LLMs are trained on extensive, proprietary datasets, making it challenging to apply certain efficiency improvement techniques that require access to the original training data~\cite{deng2020model}. This limitation not only restricts the scope of potential improvements but also raises ethical questions about transparency and the democratization of AI advancements. 2. Closed source models and lack of parameter access: Many advanced LLMs are closed source~\cite{rae2021scaling,hoffmann2022training,anil2023palm,openai2023gpt}, with restricted access to their parameters. This constraint means that efforts to improve efficiency must be conducted without deep insights into the model's internal workings, complicating the process of optimizing resource usage. The closed nature of these models also brings up ethical concerns regarding the concentration of AI capabilities and the openness of scientific research.

    \item \textbf{[Metrics]} 
    In the context of LLMs, the development of comprehensive metrics for evaluating resource efficiency faces unique challenges due to the diverse and complex nature of LLM tasks and architectures~\cite{kaddour2023challenges}. Unlike smaller models where optimizing one or two resources, such as computation or memory, might be sufficient, LLMs present a multi-objective problem requiring simultaneous optimization across multiple key resources, including computation, memory, energy, monetary cost, etc~\cite{gunantara2018review}. Therefore, a comprehensive metric for LLMs must provide a holistic view that encapsulates all these critical resources, quantifying not only the individual resource usage but also the interdependencies and trade-offs between them~\cite{menghani2021efficient}. This approach is crucial for advancing LLMs in a balanced and sustainable manner, significantly more complex than metric development for smaller models.

\end{itemize}

    % \item \textbf{The lifecycle of an LLM, which includes pre-training, fine-tuning, and inference, introduces stage-specific challenges.} Techniques that optimize resource usage during the intensive pre-training phase, where the model learns from vast amounts of data, might not be as effective or applicable during fine-tuning and inference phases. Each stage has its optimization needs and constraints, requiring tailored strategies rather than a one-size-fits-all approach.
    
    % \item \textbf{The efficiency of LLMs is not solely determined by model and algorithm design but is also critically dependent on system and hardware design.} The infrastructure underlying these models must be optimized for the specific computational demands of LLMs. This includes the design of specialized hardware that can efficiently process the massive parallel computations LLMs require and the development of systems that can facilitate the rapid and efficient communication of data across distributed networks.
    
    % \item \textbf{The inherent complexity of natural language processing adds another layer of difficulty.} LLMs must capture and generate nuanced, contextually rich language. Over-aggressive optimization techniques might streamline resource use but at the cost of the model's ability to handle the subtleties and complexities of human language. Ensuring that resource-efficiency measures do not diminish the model's linguistic capabilities is a critical and intricate challenge.

In recent years, significant research efforts have been dedicated to developing and applying resource-efficient LLMs
to address the challenges referenced earlier. There has been a wave of research proposing and deploying new strategies across various fields, although the concept of resource-efficient LLMs is relatively nascent. Most existing LLM approaches have been tailored for specific application domains; however, the underlying principles are often adaptable enough to be utilized in other areas. Nevertheless, it remains challenging to compare these resource-efficient strategies across different domains that cater to distinct communities. Furthermore, assessing the performance of resource-efficient LLMs demands intricate and specialized evaluation strategies due to their distinctive attributes, such as their multi-dimensional efficiency (e.g., computational, energy, memory usage) and the diverse outcomes they produce. To date, there is a deficiency in systematic standardization and a comprehensive summarization framework to evaluate the various methodologies proposed for resource-efficient LLMs. This lack of a cohesive summary and classification of existing methods and applications in resource-efficient LLMs poses significant issues for practitioners who need clear information on current limitations, pitfalls, unresolved questions, and promising directions for future research.

In response to these gaps, this paper seeks to offer a systematic review of the techniques, benchmarks, and evaluation metrics that contribute to the resource efficiency of LLMs. To our knowledge, this constitutes the first detailed survey explicitly devoted to resource efficiency in the context of LLMs. In the following, we outline the principal contributions of this survey:

\begin{itemize}

    \item \textbf{Comprehensive overview of resource-efficient LLM techniques:} Our paper makes a significant contribution by offering a comprehensive overview of techniques aimed at enhancing the resource efficiency of Large Language Models. This overview is extensive, covering the entire range of the LLM lifecycle. It delves into various methodologies and strategies developed in the field, focusing on how they contribute to making LLMs more resource-efficient. 
    
    \item \textbf{Systematic categorization and taxonomy of techniques by resource type:} We established a systematic categorization and taxonomy of resource-efficient LLM techniques, organized primarily by the type of resource(s) they optimize. This taxonomy simplifies the process of identifying and selecting appropriate methods based on specific resource optimization needs and provides a clear and organized framework that aids researchers and practitioners in navigating the landscape of resource-efficient LLMs.

    % \item \textbf{A systematic categorization and taxonomy of resource-efficient LLM techniques:} We offer a comprehensive classification of existing methods aimed at enhancing the resource efficiency of LLMs. These methods are organized based on the type of resource they optimize—such as computational, memory, energy, financial, and network communication—and are further categorized by their applicability in the lifecycle of LLMs, including architecture design, pre-training, fine-tuning, inference, and system design. This framework is intended to aid practitioners and researchers in pinpointing the most effective strategies for their specific resource constraints and model requirements.
    
    % \item \textbf{Comprehensive Overview of :} We introduce a taxonomy of application domains that are particularly affected by resource efficiency in LLMs. The survey underscores the practical importance and unique challenges associated with each domain, facilitating the alignment of resource optimization techniques with domain-specific objectives. This enables researchers and domain experts to cross-reference and expand their techniques to new application domains while critically evaluating their methods.
    
    \item \textbf{Standardization of evaluation metrics and datasets:} We present a standardized set of evaluation metrics and datasets tailored for assessing the resource efficiency of LLMs. This standardization facilitates consistent and fair comparisons across different models and techniques and provides a benchmark for future research in the field.
    
    \item \textbf{Identification of gaps and future research directions:}  The paper concludes with a thoughtful discussion of the current bottlenecks and unresolved challenges in creating resource-efficient LLMs. By examining the limitations of existing approaches, we shed light on potential avenues for future research. 
    
\end{itemize}

\subsection{Related work}
\label{sec: related work}

In this section, we discuss the relationship between our survey and some existing surveys on similar topics. In general, we can divide existing surveys related to the efficiency and acceleration of LLMs into the following categories: 1. fundamental overview of LLMs; 2. survey of model compression for LLMs; and 3. review of techniques of efficiency and acceleration for general deep neural networks.

\begin{itemize}
    \item \textbf{Fundamental overview of LLMs.} With the recent surge in the popularity and efficacy of LLMs, numerous review papers have surfaced, offering insights into various aspects of LLMs. Some concentrate on dissecting the fundamental components of LLMs~\cite{liu2023summary,yang2023harnessing,zhao2023survey}, while others delve into the historical context and potential applications of generative AI~\cite{zhang2023complete,cao2023comprehensive,ling2023domain}. A select few~\cite{mialon2023augmented} explore strategies for enhancing LLMs with reasoning capabilities. Nevertheless, a comprehensive review and technical taxonomy specifically focused on the specialization of LLM domains remains an unaddressed gap in the current literature.
    
    \item \textbf{Survey of compression and acceleration for LLMs.} Transformer-based language models have achieved huge success, however, the computational and memory cost remains a big concern despite the superior performance. There have been several survey papers on how to compress and accelerate large language models. For example, some discuss how to accelerate the inference of LLMs~\cite{xu2023survey,zhu2023survey,chitty2023survey}, by focusing on model compression techniques. In addition, a select few~\cite{tay2020efficient,fournier2023practical} explore more efficient and lightweight architecture designs for transformers, which are the backbone of modern LLMs. Furthermore, some works discuss the efficient training of LLMs~\cite{zhuang2023survey}. However, those existing surveys either lack comprehensiveness or are not up-to-date, especially considering the large number of papers published after the birth of ChatGPT, which marks the beginning of the LLMs era.
    
    \item \textbf{Review of efficient deep neural networks.} How to achieve efficient design or accelerate the computation of deep neural networks (DNNs) has long been a popular research direction, and there have been a couple of survey papers on this topic. Some works focus on the model compression and acceleration of DNNs~\cite{cheng2017survey,long2019survey}. A few others discuss the hardware design and optimization for DNNs~\cite{capra2020updated,dhilleswararao2022efficient}. However, due to the very large model size and the special architecture of transformers, there is a big gap in directly applying those techniques for DNNs onto the LLMs.
\end{itemize}

\subsection{Outline}
\label{sec: outline}

The remainder of this survey is structured as follows, offering a detailed exploration of resource-efficient LLMs:

\begin{itemize}
    \item Section~\ref{sec: preliminary and taxonomy} \textit{Preliminary and taxonomy}: This section sets the foundation by introducing the fundamental concepts behind transformers and pre-trained LLMs. It establishes a comprehensive taxonomy of resources essential for LLMs, such as computation, memory, energy, money, and network communication. This taxonomy serves as a guiding framework for the entire survey, outlining the key areas of focus for improving resource efficiency in LLMs.

    \item Section~\ref{sec: architecture design} \textit{LLM architecture design}: This section delves into the latest developments in LLM architecture, emphasizing designs that enhance resource efficiency. It discusses both efficient transformer architectures, which optimize traditional transformer models, and non-transformer architectures which propose alternative structures for resource optimization.

    \item Section~\ref{sec: pretrain} \textit{LLM pre-training}: This section explores the various pre-training techniques for LLMs, highlighting how they contribute to resource efficiency. Key areas such as memory efficiency, data efficiency, and innovative training pipeline designs are examined, illustrating how each technique impacts the overall resource utilization during the pre-training phase.

    \item Section~\ref{sec: finetune} \textit{LLM fine-tuning}: This section covers the fine-tuning phase of LLMs, focusing on methods that enhance resource efficiency. It includes detailed discussions on parameter-efficient fine-tuning, which minimizes parameter updates; and full-parameter fine-tuning, which optimizes the entire parameter set.

    \item Section~\ref{sec: inference} \textit{LLM inference}: Here, we analyze various inference techniques that improve resource efficiency in LLMs. The section features discussions on static methods including pruning, quantization, knowledge distillation, low-rank approximation, etc. In addition, we also discuss dynamic methods such as dynamic inference, which adapts computational resources in real time, and token parallelism, which optimizes processing at the token level to enhance efficiency during the inference stage.

    \item Section~\ref{sec: system design} \textit{System design}: This section addresses system-level strategies for resource-efficient LLMs, encompassing support infrastructure, which focuses on leveraging specialized systems for efficiency, and deployment optimization, which involves strategies for deploying LLMs in a resource-conscious manner.

    \item Section~\ref{sec: application} \textit{Technique categorization by resources}: In this section, we evaluate the effectiveness of various resource efficiency techniques. The discussion revolves around real-world applications and how different methods fare in practical scenarios, providing a bridge between theory and application.

    \item Section~\ref{sec: dataset and metrics} \textit{Benchmark and evaluation metrics}: This section presents the benchmarks and metrics used for evaluating the resource efficiency of LLMs. It highlights the importance of standardized evaluation criteria in assessing the effectiveness of various techniques and models.

    \item Section~\ref{sec: challenges} \textit{Open Challenges and future directions}: Here, we identify the existing challenges and potential future research directions in the field of resource-efficient LLMs. This section is crucial for understanding the current gaps in the field and where future efforts may be most beneficial.

    \item Section~\ref{sec: conclusion} \textit{Conclusion}: The survey concludes with a summary of the key findings and insights presented, encapsulating the core takeaways from the exploration of resource efficiency in LLMs.
\end{itemize}

\section{Preliminary and taxonomy}
\label{sec: preliminary and taxonomy}

In this section, we first provide some preliminaries of this survey, including some introduction about transformers and LLMs. Then, we introduce our proposed taxonomy of the techniques for the efficiency and acceleration of LLMs.

\subsection{Preliminaries}
\label{sec: preliminaries}

\subsubsection{Transformer model}

The Transformer model stands as a pivotal milestone in the evolution of deep learning, particularly in the realm of natural language processing (NLP). Introduced by~\cite{vaswani2017attention}, the Transformer model represents a groundbreaking departure from conventional sequence-to-sequence models, offering an innovative solution to the challenges of capturing long-range dependencies in sequences.

\begin{itemize}
    \item \emph{Embedding Layers.} The embedding layer is the foundational component of a Transformer model, serving as the initial step in transforming raw input data into a format that can be effectively processed. It maps discrete tokens, such as words or subwords, into continuous vector representations, often referred to as word embeddings. These embeddings capture semantic relationships between words and enable the model to understand the meaning of each token. 
    \item \emph{Positional Encoding.} Unlike recurrent neural networks (RNNs) or convolutional neural networks (CNNs), Transformers do not inherently possess knowledge of the order or position of tokens in a sequence. To address this limitation, positional encodings are introduced. These encodings are added to the word embeddings and provide the model with information about the position of each token within the sequence. Typically, sinusoidal functions are used to generate these positional encodings, ensuring that the model can capture sequential dependencies without relying on recurrence or convolution.
    \item \emph{Self-Attention.} Self-attention is the cornerstone of the Transformer architecture and allows the model to weigh the importance of different words in the input sequence when making predictions for a particular word. It computes a weighted sum of all input words, where the weights are determined dynamically based on the similarity between words. The self-attention mechanism is computed using a weighted sum over all words in the sequence, and the weights are determined by the dot product of query, key, and value vectors:
    \begin{equation}
        \text{Attention}(Q, K, V) = \text{softmax}\left(\frac{{QK^T}}{\sqrt{d_k}}\right) V.
        \label{eq: self-attention}
    \end{equation}
    Here, \(Q\), \(K\), and \(V\) are the query, key, and value matrices, respectively, and \(d_k\) is the dimension of the key vectors. This mechanism allows the model to focus on relevant information within the input sequence.
    \item \emph{Multi-Head (Self-)Attention.}  Multi-head self-attention extends the self-attention mechanism by performing it multiple times in parallel, with different sets of learned parameters. This allows the model to capture different types of relationships and dependencies in the input sequence, providing a richer representation. Mathematically, multi-head self-attention involves computing multiple sets of query, key, and value matrices, and then concatenating the results from each head:
    \begin{equation}
        \text{MultiHead}(Q, K, V) = \text{Concat}\big(\text{head}_1, \text{head}_2, \ldots, \text{head}_h\big)W^O,
        \label{eq: multi-head attention}
    \end{equation}
    where \(\text{head}_i = \text{Attention}(QW_i^Q, KW_i^K, VW_i^V)\) represents the output of the \(i\)-th attention head, and \(W^O\) is a learned linear transformation. Multi-head self-attention enhances the model's ability to capture both local and global dependencies in the data.
\end{itemize}

In summary, the Transformer architecture consists of these key components, each playing a crucial role in enabling the model to understand and generate sequences effectively, making it a powerful tool in natural language processing and other sequence-to-sequence tasks.

\subsubsection{Large Language Models (LLMs)}

Pre-trained Language Models (PLMs) constitute a type of neural network that has been trained on extensive collections of text data. Their purpose is to acquire knowledge of linguistic patterns, structures, and semantics inherent in the language. 
In the context of LLMs, the input comprises a text sequence that serves as the context for comprehension and processing. Often, a prompt or additional sentence is included to clarify the task. These prompts are tailored to the specific NLP task at hand, providing a premise or task explanation. For example, in text summarization, a prompt like ``Summarize the key points in the following passage:" can be placed before the input passage. The output is the generated text sequence or prediction responding to the input, e.g., the summarized key points of the provided passage. In some cases, post-processing steps such as token decoding or label extraction may be necessary for the final presentation.

LLMs typically follow the architectural designs of PLMs and come in three primary flavors: encoder-only, encoder-decoder, and decoder-only architectures. Here's an overview of these LLM architectures and their distinctions:

\begin{itemize}
    \item \emph{Encoder-only Language Models.} These models process input text to create vector representations without an explicit decoding phase for generating new text. Instead, they transform and embed text into a high-dimensional space. Encoder-only models are primarily designed to capture and understand patterns and semantics in the input data. They find extensive use in tasks such as text classification, sentiment analysis, and clustering. A notable example is BERT~\cite{devlin2018bert}, which extracts context-rich embeddings for downstream tasks through pre-training on a masked language modeling objective.
    \item \emph{Encoder-Decoder Language Models.} These models consist of an encoder that processes input text into vector representations and a decoder that generates output text based on these representations. They employ cross-entropy loss as the objective function, comparing the actual and predicted target sequences. Encoder-Decoder LLMs are often used for sequence-to-sequence tasks such as machine translation and summarization. T5~\cite{raffel2020exploring} is a notable example of this architecture.
    \item \emph{Decoder-only Language Models.} Examples like GPT~\cite{brown2020language} are autoregressive language models that generate the next word in a sequence based on previous words. They map a sequence of tokens to a vector representation and generate contextually relevant content autoregressively, calculating the probability of the next token based on the context. This autoregressive approach is particularly suitable for text-generation tasks.
\end{itemize}

In summary, LLMs and their variants play a pivotal role in natural language processing tasks by leveraging pre-training on vast text corpora to facilitate a wide range of language understanding and generation tasks.

\subsection{Proposed taxonomy}
\label{sec: taxonomy}

\subsubsection{Taxonomy of key resources involved with using LLMs.}
The taxonomy for resource efficiency in LLMs encompasses five key domains: computation, memory, energy, money, and network communication. Each domain addresses a distinct aspect of resource utilization:
\begin{itemize}
    \item \textbf{Computation:} This involves the processing power required for tasks such as training, fine-tuning, and executing LLMs. Evaluating computational efficiency includes considering the number of operations (like floating-point operations), the efficiency of algorithms, and the utilization of processing units like GPUs or TPUs. It is crucial to explore how to maximize output while minimizing computational requirements.
    
    \item \textbf{Memory:} Memory efficiency pertains to the amount of RAM and storage needed. LLMs, especially those with billions of parameters, require significant memory for storing the model weights and for processing large datasets during training and inference. Optimizing data structures, employing techniques like model pruning, and exploring memory-efficient architectures are key strategies here.
    
    \item \textbf{Energy:} This resource refers to the electrical power consumed during the model's lifecycle. Given the environmental impact and operating costs, energy efficiency is vital. It includes strategies for reducing power consumption, such as optimizing hardware utilization, using energy-efficient hardware, and implementing algorithms that require less computational power.
    
    \item \textbf{Money:} Financial resources are a crucial consideration, especially for smaller organizations and researchers. This includes the cost of hardware acquisition, electricity for running the models, and potential cloud computing expenses. Finding ways to make LLMs accessible and viable for a broader range of users without significant financial investment is another key challenge.
    
    \item \textbf{Network communication:} For distributed training and cloud-based deployment, network bandwidth and latency become significant. Efficient network communication means reducing the amount of data that needs to be transferred between nodes in a distributed system or between the cloud and end-users, which can greatly affect training time and responsiveness in real-time applications.
\end{itemize}

\forestset{
  L1/.style={fill=white, draw=black, rotate=90, text width=4cm},
  L2/.style={fill=white, draw=black, text width=2.9cm},
  L3/.style={fill=white, draw=black, text width=4cm},
  L4/.style={fill=white, draw=black, text width=3.9cm},
}
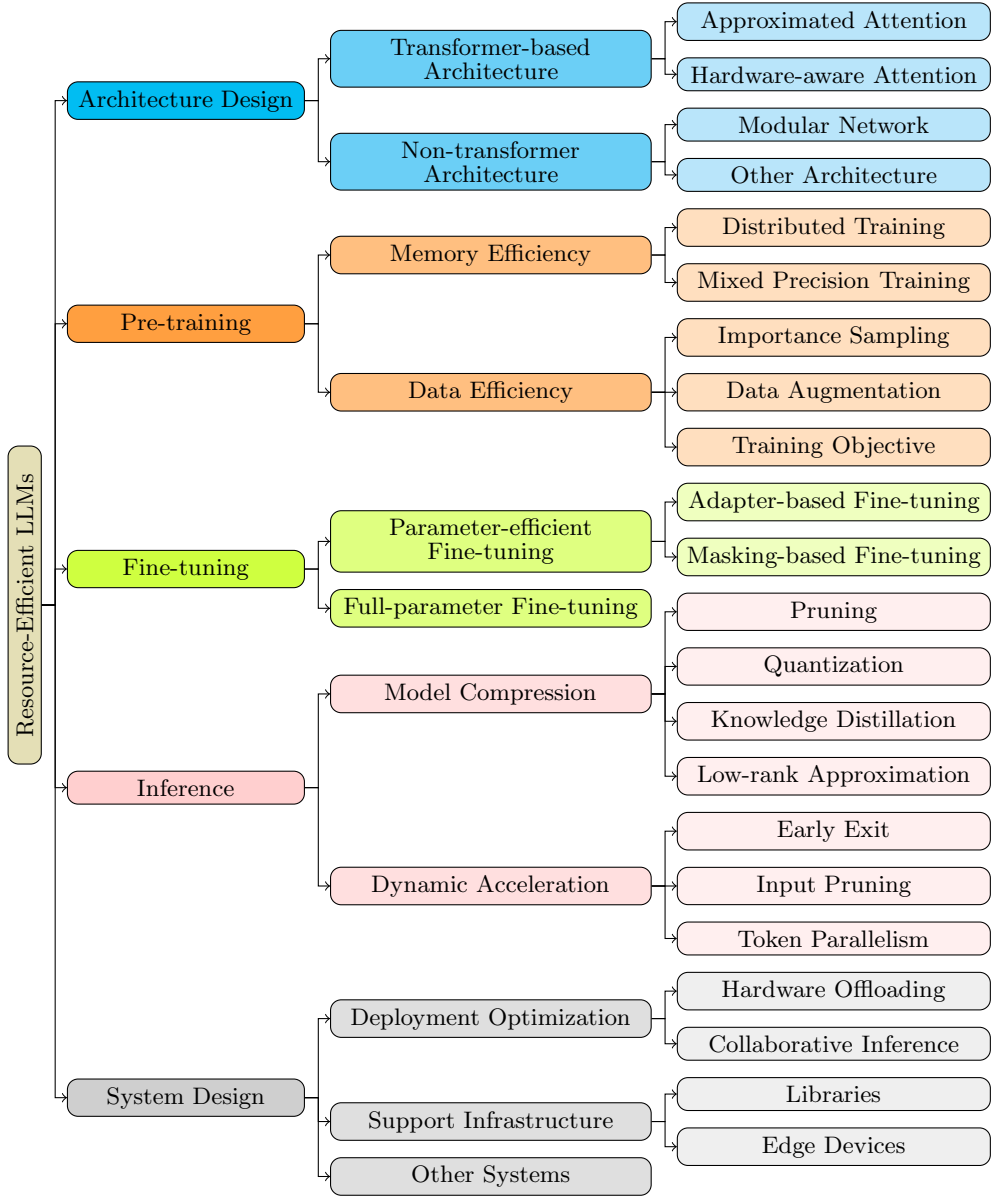
\begin{figure}[t!]
\scriptsize
\begin{forest}
for tree = {draw,rounded corners,grow'=0,text width=2cm, text centered,edge+={->}}, forked edges
% where level<=1{rot}{},
[,phantom,
    [,tier = a,draw=none, text width=0cm, for tree={color=gray, text centered, no edge, draw=none}]        
    [Resource-Efficient LLMs, L1, fill=olive!25
      [Architecture Design, L2, fill=cyan!75, tier=b,
        [Transformer-based Architecture, L3, fill=cyan!50, tier=c
          [Approximated Attention, L4, fill=cyan!25, tier=d]
          [Hardware-aware Attention, L4, fill=cyan!25, tier=d]
        ]
        [Non-transformer Architecture, L3, fill=cyan!50, tier=c
          [Modular Network, L4, fill=cyan!25, tier=d]
          [Other Architecture, L4, fill=cyan!25, tier=d]
        ]
      ]
      [Pre-training, L2, fill=orange!75, tier=b,
        [Memory Efficiency, L3, fill=orange!50, tier=c,
            [Distributed Training, L4, fill=orange!25, tier=d]
            [Mixed Precision Training, L4, fill=orange!25, tier=d]
        ]
        [Data Efficiency, L3, fill=orange!50, tier=c,
            [Importance Sampling, L4, fill=orange!25, tier=d]
            [Data Augmentation, L4, fill=orange!25, tier=d]
            [Training Objective, L4, fill=orange!25, tier=d]
        ]
      ]
      [Fine-tuning, L2, fill=lime!75, tier=b,
        [Parameter-efficient Fine-tuning, L3, fill=lime!50, tier=c,
            [Adapter-based Fine-tuning, L4, fill=lime!25, tier=d]
            [Masking-based Fine-tuning, L4, fill=lime!25, tier=d]
        ]
        [Full-parameter Fine-tuning, L3, fill=lime!50, tier=c
            % [?, L4, fill=lime!25, tier=d]
        ]
        % [?, L3, fill=lime!50, tier=c]
      ]
      [Inference, L2, fill=pink!75, tier=b,
        [Model Compression, L3, fill=pink!50, tier=c,
            [Pruning, L4, fill=pink!25, tier=d]
            [Quantization, L4, fill=pink!25, tier=d]
            [Knowledge Distillation, L4, fill=pink!25, tier=d]
            [Low-rank Approximation, L4, fill=pink!25, tier=d]
        ]
        [Dynamic Acceleration, L3, fill=pink!50, tier=c,
            [Early Exit, L4, fill=pink!25, tier=d]
            [Input Pruning, L4, fill=pink!25, tier=d]
            [Token Parallelism, L4, fill=pink!25, tier=d]
        ]
        % [System Design, L3, fill=pink!50, tier=c,
        %     [Hardware Offloading, L4, fill=pink!25, tier=d]
        %     [Collaborative Inference, L4, fill=pink!25, tier=d]
        % ]
      ]
      % [Hardware Optimization, L2, fill=lightgray!75, tier=b,
      %   [Libraries, L3, fill=lightgray!50, tier=c]
      %   [Edge Devices, L3, fill=lightgray!50, tier=c]
      % ]
      [System Design, L2, fill=lightgray!75, tier=b,
        [Deployment Optimization, L3, fill=lightgray!50, tier=c
            [Hardware Offloading, L4, fill=lightgray!25, tier=d]
            [Collaborative Inference, L4, fill=lightgray!25, tier=d]
        ]
        [Support Infrastructure, L3, fill=lightgray!50, tier=c
            [Libraries, L4, fill=lightgray!25, tier=d]
            [Edge Devices, L4, fill=lightgray!25, tier=d]
        ]
        [Other Systems, L3, fill=lightgray!50, tier=c]
      ]
    ]
]
\end{forest}
\normalsize
\caption{A taxonomy of techniques for achieving resource-efficient LLMs.}
\label{fig: technical taxonomy}
\end{figure}

\subsubsection{Taxonomy of techniques for resource-efficient LLMs}

As delineated in Figure~\ref{fig: technical taxonomy}, our survey paper introduces a structured taxonomy that categorizes techniques for enhancing the resource efficiency of LLMs into clear, defined tiers. We propose five principal categories: Architecture Design, Pre-training, Fine-tuning, Inference, and System Design. Each of these is selected for its integral role in the lifecycle of efficient LLM development and deployment.

\begin{itemize}

    \item \textbf{Architecture design.} This category examines the structural foundations of LLMs, branching into Transformer-based and Non-transformer Architectures. These classifications are intended to highlight architectural variations and innovations crucial for the models' efficiency and efficacy.

    \item \textbf{Pre-training.} This category inspects the preliminary phases of LLM development, including Memory Efficiency and Data Efficiency. It underscores the importance of the pre-training environment and strategies that significantly affect the models' future performance and resource utilization.

    \item \textbf{Fine-tuning.} Addressing the optimization of pre-trained models, this category is organized into Parameter-efficient Fine-tuning and Full-parameter Fine-tuning. These subdivisions represent the range of techniques that refine models for particular tasks or enhance their overall functionality.

    \item \textbf{Inference.} During the operational stage, various strategies under the Inference category, such as Model Compression and Dynamic Acceleration, are employed. This classification acknowledges the diverse tactics applied at the model inference phase, impacting efficiency and performance distinctly.

    \item \textbf{System design.} Focusing on system-level considerations, this category covers Deployment Optimization and Support Infrastructure, among others. It explores hardware and system optimizations that are essential for improving the practical performance of LLMs.
    
\end{itemize}

Through this taxonomy, we aim to facilitate a structured and nuanced understanding of the diverse methodologies and strategies employed in the quest for enhanced efficiency and acceleration of LLMs, providing a holistic view of the current research landscape.

\section{LLM architecture design}
\label{sec: architecture design}

This section explores the advancements in architecture design for LLMs, specifically focusing on enhancing the efficiency of Transformer models. We examine various strategies aimed at reducing computational and memory demands, crucial for the practical deployment of LLMs. The discussion includes innovative approaches like Reformer, Linear Transformer, AFT, and KDEformer, each presenting unique solutions to optimize processing speed and resource usage. Additionally, we touch upon hardware-optimized attention mechanisms and alternative non-transformer architectures, highlighting their contributions to the evolving landscape of efficient LLM design.

\subsection{Efficient transformer architecture}
\label{sec: efficient transformer architecture}

Efficient transformers focus on creating neural network architectures that are optimized for enhanced throughput. The attention layer significantly influences the processing speed of transformers, which contributes a lot to the throughput. 

\subsubsection{Approximate attention.} One stream of works focuses on designing attention operators with approximation techniques to achieve \textbf{less time complexity and/or less memory complexity}. 
In the classic Transformer, the time complexity of the self-attention operator is $\mathcal{O}(T^{2}d)$, and the memory complexity is $\mathcal{O}(T^2)$. Here $T, d$ denote the sequence length and hidden feature dimension, respectively.
Reformer~\cite{Kitaev2020Reformer} replace dot-product attention by proposed locality-sensitive hashing attention, which leads to $\mathcal{O}(Td \log T)$ time complexity and $\mathcal{O}(T\log T)$ memory complexity.
Linear Transformer~\cite{katharopoulos2020linearTrans} expresses the self-attention as a linear dot-product of kernel feature maps and utilizes the associativity property of matrix products to reduce the complexity term from $T^2$ to $T$, thus achieving $\mathcal{O}(Td^{2})$ time complexity and $\mathcal{O}(Td+d^2)$ memory complexity. 
EfficientAttention~\cite{shen2021efficient} proposes an approximated attention operation by switching the $\mathbf{QKV}$ multiplication order from $\mathbf{(QK^T)V}$ to $\mathbf{Q(K^TV)}$, which leads to more efficient $\mathcal{O}(T^{2}d)$ time complexity and same $\mathcal{O}(Td+d^2)$ memory complexity when $d < T$. 
AFT~\cite{zhai2021aft} proposes an extremely efficient variant called AFT-simple, which achieves linear complexity in both time ($\mathcal{O}(Td)$) and memory ($\mathcal{O}(Td)$). In an AFT layer, the key and value are first combined with a set of learned position biases ($s<T$), so that the multiplication between the key-value and query is in an element-wise manner. The introduced learned position bias $s$ can be eliminated in the AFT-simple variant (\textit{i.e.} no position bias is learned) so that AFT-simple completely gets rid of the need for dot products operations. 
Memory efficient attention~\cite{rabe2021memEffAttn} presents a practical implementation for self-attention that requires only $\mathcal{O}(d \log T)$ memory with the same time complexity $\mathcal{O}(T^2d)$. The core idea behind memory efficient attention is similar to ``lazy softmax''~\cite{jang2019mnnfast} where the denominator of the softmax for the dot product of queries and keys can be calculated in the later stage.
KDEFormer~\cite{zandieh2023kdeformer} suggests reducing the denominator of the softmax function to a variant of the kernel density estimation (KDE) problem, and an efficient KDE solver can be further utilized to accelerate the multiplication of the attention matrix with the value matrix. The trick is based on reducing the number of columns of the attention matrix (referred to as $A:=exp(QK^\top/\sqrt{d})$) using importance sampling. KDEFormer delivers a $\mathcal{O}(Tmd)$ time complexity and $\mathcal{O}(Tm)$ memory complexity, where $m<T$ is a small number.
MEGA~\cite{ma2023mega} introduces a moving average equipped gated attention mechanism to solve the weak inductive bias and quadratic computational complexity. MEGA offers $\mathcal{O}(cTd)$ time complexity and $\mathcal{O}(cTd)$ memory complexity with a theoretical grounding, where $c < T$ is MEGA's chunk size of quadratic attention. 
LoMA~\cite{wang2024loma} introduces a method for losslessly compressing the memory of transformer-based language models, allowing for a substantial increase in contextual length without altering the model architecture. By segmenting inputs into reading, memory, and repetition areas, LoMA utilizes a bidirectional attention mask within the memory area to preserve information. The approach achieves up to a 4:1 compression ratio, maintaining the model's generative capability through fine-tuning and enabling efficient long-text handling with minimal data requirements.
BiPE~\cite{he2024two} introduces a bilevel positional encoding approach that combines intra-segment and inter-segment encodings to enhance length extrapolation capabilities in transformer models. This design disentangles positional information within and between segments, allowing for more efficient encoding. BiPE achieves superior length extrapolation performance with a theoretical advantage, delivering a time complexity of $\mathcal{O}(Td)$ and memory complexity of $\mathcal{O}(Td)$ across diverse tasks.
Simple Linear Attention~\cite{arora2024simplelinearattentionlanguage} combines sliding window and linear attention mechanisms, offering a solution to the recall-throughput tradeoff by balancing memory consumption and token recall. It delivers $\mathcal{O}(Td^2)$ time complexity and utilizes a hardware-optimized IO-aware algorithm, achieving up to 24× higher throughput than FlashAttention-2, making it a highly efficient architecture for language generation.
Cluster-wise Graph Transformer (Cluster-GT) introduces the Node-to-Cluster Attention (N2C-Attn) mechanism~\cite{huang2024clusterGT}, leveraging Multiple Kernel Learning in a kernelized attention framework to capture node and cluster-level information without compressing clusters into single embeddings, achieving linear time complexity and excelling in graph-level tasks by integrating dual-granularity feature maps through an efficient cluster-wise message-passing architecture.
SageAttention~\cite{zhang2024sageattention} introduces a novel quantization method specifically for the attention mechanism, achieving approximately 2.1× and 2.7× higher OPS than FlashAttention2 and xformers, respectively, while maintaining superior accuracy to FlashAttention3, thus enabling efficient model inference with minimal end-to-end performance loss across large language, image, and video generation models.
Local Attention Mechanism (LAM)~\cite{aguileramartos2024localattentionmechanismboosting} leverages the continuity of time series data to reduce attention computations, achieving $\mathcal{O}(n \log n)$ time and memory complexity, significantly improving upon traditional $\mathcal{O}(n^2)$ complexity, and demonstrates superior performance in long-horizon forecasting, surpassing state-of-the-art models and addressing the need for new evaluation datasets in time series forecasting.
The proposed Long LoRA Pereceiver (LLP)~\cite{mahmood2024enhancedcomputationallyefficientlong} framework builds upon the PerceiverAR architecture to effectively cut down the quadratic complexity of traditional Transformer-based attention, achieving semi-linear time complexity, and demonstrates notable improvements over existing state-of-the-art models, positioning LLP as a compelling and efficient core component for next-generation Large Language Models.
Signformer~\cite{yang2024signformerneededgeai} introduces a from-scratch transformer pipeline with novel convolution-attention integration, achieving substantial parametric (467-1807x) and computational efficiency over contemporary SOTAs, attaining near-LLM-level performance, securing the 2nd place on the leaderboard, and demonstrating the feasibility of sustainable, edge-deployable sign language translation without reliance on large pretrained models or extensive datasets

\begin{table}[t!]
\small
    \centering
    \begin{tabular}{c|cc}
    \toprule
    Approach & Time Complexity & Memory Complexity\\
    \midrule
    Transformer~\cite{vaswani2017attention} & $\mathcal{O}(T^2d)$& $\mathcal{O}(T^2+Td)$ \\
    \hline
    Reformer~\cite{Kitaev2020Reformer}& $\mathcal{O}(Td \log T)$ & $\mathcal{O}(T\log T+Td)$\\
    Linear Transformer~\cite{katharopoulos2020linearTrans} & $\mathcal{O}(Td^{2})$ & $\mathcal{O}(Td+d^2)$ \\
    Efficient Attention~\cite{shen2021efficient} & $\mathcal{O}(T^{2}d)$ & $\mathcal{O}(Td+d^2)$ \\
    AFT~\cite{zhai2021aft} & $\mathcal{O}(Td)$ & $\mathcal{O}(Td)$\\
    Memory Efficient Attention~\cite{rabe2021memEffAttn} & $\mathcal{O}(T^2d)$ & $\mathcal{O}(d \log T)$\\
    KDEformer~\cite{zandieh2023kdeformer} & $\mathcal{O}(mTd)$ & $\mathcal{O}(mT)$\\
    MEGA~\cite{ma2023mega} & $\mathcal{O}(cTd)$ & $\mathcal{O}(cTd)$\\
    Simple Linear Attention~\cite{arora2024simplelinearattentionlanguage} & $\mathcal{O}(Td^2)$ &  $\mathcal{O}(Td^2)$\\
    \hline 
    RWKV~\cite{peng2023rwkv} & $\mathcal{O}(Td)$ & $\mathcal{O}(d)$ \\
    \bottomrule
    \end{tabular}
    \caption{Overview of time complexity and memory complexity improvements for selected approaches over classical Transformer. Here, $T, d$ denote the sequence length and hidden feature dimension, respectively. $m$ used in KDEformer denotes $m$ sampled columns of attention matrix. $c$ used in MEGA denotes its chunk size of quadratic attention. }
    \label{tab:attn_complexity}
\vspace{-3mm}
\end{table}

\subsubsection{Hardware optimized attention.} There is another stream of works focusing on \textbf{hardware efficient attention operator}. 
Starting from 2021, many works (LightSeq~\cite{wang2021lightseq}, Faster Transformer~\cite{FasterTransformer}, xFormers~\cite{xFormers2022}) have been focused on optimizing CUDA implementation of attentions and transformer layers including kernels fusion, gemm optimization, \textit{etc.}
FlashAttention~\cite{dao2022flashattention} introduces an IO-aware precise attention algorithm that employs tiling to minimize the volume of memory reads/writes between the high bandwidth memory of the GPU and the on-chip SRAM. Building on this, FlashAttention-2~\cite{dao2023flashattention2} further refines FlashAttention by addressing the suboptimal work partitioning concern.
vLLM~\cite{kwon2023vLLM} proposes a novel attention algorithm, PagedAttetion, that mainly optimizes the virtual memory and paging techniques in operating systems.
MobileLLM~\cite{liu2024mobilellm} introduces a deep-and-thin model structure optimized for on-device use cases, leveraging embedding sharing and grouped-query attention to enhance performance. With innovations such as block-wise weight sharing, MobileLLM achieves state-of-the-art results for sub-billion parameter models, offering $\mathcal{O}(Td)$ time complexity and maintaining model efficiency even in memory-constrained environments.

\subsection{Non-transformer architecture}
\label{sec: efficient non-transformer}

While Transformers, with their self-attention mechanisms, have dominated the field of language modeling, alternative architectures have emerged to tackle various challenges or provide different advantages.

\subsubsection{Modular network}
Modular Network (also called the Mixture of Experts (MoEs))~\cite{shazeer2016outrageously,lepikhin2020gshard} technique is a machine learning approach that combines multiple specialized models, known as experts, to solve complex tasks more effectively. As we know, a single dense LLM itself contains billions of parameters, which is extremely difficult to be further scaled into larger parameter sizes. MoE provides a solution to enabling LLM's parameter size to grow from hundreds of billions into trillions, by sparse routing (also mentioned as \textit{sparse activation}, \textit{sparse gating}). During training, multiple individual expert LLMs and a routing function are trained simultaneously. The learned routing function allows the MoE system to select a subset of experts according to the input, thus reducing computational and memory requirements. 
Switch Transformer~\cite{fedus2022switch} follows the principle of ``increasing the parameter count while keeping the floating point operations (FLOPs) per example constant''. Switch Transformer achieves this by replacing the original dense feed-forward network layer in the Transformer with a sparse Switch feed-forward layer. The Switch layer is essentially similar to ~\cite{shazeer2016outrageously} architecture but the authors simplify the number of selected experts to 1. 
GLaM~\cite{du2022glam} has 1.2T parameters which is ~7X larger than GPT-3 and requires half of the FLOPs for inference. GLaM is implemented with 64 experts per MoE layer where each input token only activates 96.6B (8\% of 1.2T) parameters. Different from the Switch Transformer, it contains an MoE layer interleaved with a traditional Transformer layer in each block of GLaM. 
Concurrent works include heterogeneous MoEs~\cite{zhou2022mixture}, MoE-LM~\cite{artetxe2022efficient}, Unifed Routing Network~\cite{clark2022unified}.

\subsubsection{Other architecture}
Researchers have explored more novel dense architectures that are different from the transformers. 
Inspired by AFT introduced in Section~\ref{sec: efficient transformer architecture}, RWKV~\cite{peng2023rwkv} combines the efficient parallelizable training of Transformers with the efficient inference of RNNs. The key idea behind RWKV is to leverage a linear attention mechanism so that the proposed model can be formulated as a transformer during training and an RNN during inference. 
On the other hand, \cite{malach2023auto} explores the potential of Multi-Layer Perceptrons trained with the same next-token prediction to achieve non-trivial performance on text generation and arithmetic tasks. The authors also supply rich theory analysis to connect and compare their proposed architecture to existing transformer-based architectures, and they argue that the power of LLM can mostly be attributed to the large-scale auto-regressive next-token training scheme. 
Hyena~\cite{poli2023hyena} proposes a subquadratic drop-in replacement for attention constructed by interleaving implicitly parametrized long convolutions
and data-controlled gating.
Monarch Mixer~\cite{fu2023monarch} utilizes a simple yet efficient subquadratic \emph{generalized matrix multiply algorithms} based architecture.
Mamba~\cite{gu2023mamba} integrates \emph{selective state space} models into a simple end-to-end neural network architecture without attention blocks, achieving competitive modeling power of Transformers while scaling linearly in sequence length.
YOCO~\cite{sun2024you} introduces a memory-efficient decoder-decoder architecture that caches key-value pairs once, enabling sublinear scaling in GPU memory consumption and substantially reducing prefilling latency across long-context language models.
MatMul-free LM~\cite{zhu2024matmul} eliminates costly matrix multiplication operations by leveraging ternary weights and element-wise Hadamard products, achieving substantial reductions in memory usage and training latency while scaling to billion-parameter language models.
RWKV-edge~\cite{choe2024rwkvedgedeeplycompressedrwkv}, including low-rank approximation, sparsity predictors, and clustering head, effectively reduce RWKV model size by 4.95–3.8x with a minimal 2.95pp drop in accuracy, facilitating the practical deployment of RNN-based LLMs on resource-constrained devices and demonstrating the viability of efficient, high-performing large language models in embedded environments.
\section{LLM pre-training}
\label{sec: pretrain}

For large-scale LLMs like GPT-4, efficient pre-training is pivotal due to their extensive size and complexity. This efficiency transcends mere speed, focusing on optimal utilization of computational resources and innovative data management. Combining advanced hardware, such as GPUs and TPUs, with techniques like data and model parallelism, the pre-training process is tailored to be resource-efficient. Additionally, strategies like selective data sampling, model pruning, and quantization play a crucial role in minimizing data and memory requirements. These methods collectively contribute to not only accelerating the training process but also ensuring sustainable and cost-effective development of advanced LLMs.

\subsection{Memory efficiency}
\label{sec: memory efficiency}

\subsubsection{Distributed training}
\label{sec: distributed training}
Distributed training proves to be a highly effective approach for accelerating model training, particularly for machine learning tasks that exceed the memory capacity of a single accelerator, such as GPU, TPU, and more. In distributed training, the task of training the model is divided and allocated to multiple working nodes. These nodes concurrently execute local training tasks and collectively contribute to developing the original task.

\noindent\textbf{Data parallelism.} Data parallelism (DP) is the most straightforward approach for distributed training and has been inherently supported by famous machine learning frameworks like TensorFlow and PyTorch. In the paradigm of data parallelism, the initial dataset is divided into multiple partitions, and different data partitions are trained in parallel by multiple accelerators. However, DP has memory redundancies across all data partitions, and model states including model parameters, gradients, and optimizer states are required by each data partition. Given the substantial size of LLMs, applying DP to LLMs in a naive manner is impractical. To end this, ZeRO~\cite{rajbhandari2020zero}, PaLM~\cite{chowdhery2023palm} and Fairscale~\cite{baines2021fairscale} introduce approaches for enhancing the efficiency of memory utilization in DP. Instead of duplicating the entire model states, these techniques suggest partitioning them. Each data partition stores a portion of the model states and can retrieve additional states from other data partitions with a dynamic communication schedule when necessary.

\noindent\textbf{Model parallelism.} Model parallelism (MP) is a kind of distributed training method that aims to minimize a model's memory footprint by spreading its layers or tensors across multiple accelerators, while DP primarily concerns data partitioning. Based on the partition levels, model parallelism can be categorized into two main types: tensor model parallelism (TMP) and pipeline model parallelism (PMP).

In the context of TMP, tensors can be split along their rows or columns, enabling concurrent execution of matrix multiplication operations across all split parts. Megatron-lm~\cite{shoeybi2019megatron} employs parallelization techniques for matrix multiplication operations within both the multi-layer perceptron (MLP) and the self-attention block of the transformer layer. In the case of the MLP, the weight matrix is divided along its columns, while for the self-attention block, the Query, Key, and Value parameters are split in a column-parallel fashion. Alternatively, Mesh-tensorflow~\cite{shazeer2018mesh} split the units in the hidden layer to achieve tensor MP.

In the case of PMP, a model is divided into multiple layer groups, and each accelerator is responsible for handling one of these groups. To minimize inter-accelerator communication, these groups typically consist of consecutive layers. While naively implementing PMP can reduce the memory demands on each accelerator, it is important to note that, due to layer dependencies, most accelerators are idle at any given time, with only one in active operation. To enhance the resource utilization, GPipe~\cite{huang2019gpipe} and PipeDream~\cite{narayanan2019pipedream} adopt an approach where a batch is divided into smaller micro-batches. PMP is then executed independently on each micro-batch, and gradient updates occur asynchronously across these micro-batches. BPipe~\cite{kim2023bpipe} aims to achieve memory balance among accelerators during the training of PMP by transferring intermediate activations. Alpa~\cite{zheng2022alpa} proposes a model-parallel training system for large deep-learning models. It has the capability to automatically generate parallel execution plans that encompass data, operator, and pipeline parallelisms. MegaScale~\cite{jiang2024megascale} introduces a scalable training system leveraging 3D parallelism and in-depth observability, achieving high training efficiency and stability across over 10,000 GPUs. ProTrain~\cite{yang2024protrain} executes PMP independently on each micro-batch, allowing asynchronous gradient updates across these batches, thus enhancing throughput efficiency.

\subsubsection{Mixed precision training} 
\label{sec: mixed precision training}
Mixed precision training is a technique used to accelerate the training of deep learning models by using both 16-bit and 32-bit floating-point types (as opposed to just using 32-bit or 64-bit throughout the training process such as BERT~\citep{devlin2018bert}). This approach has gained popularity, especially in the training of large language models, where computational cost can be a significant barrier. Recently, to pre-train extremely large language models, some works borrowed 16-bit floating-point numbers~\cite{micikevicius2017mixed} which largely reduced memory consumption compared with 32-bit or 64-bit. To mitigate the performance degradation caused by the quantization with 16-bit floating-point numbers, Scao et al.~\citep{scao2022bloom} proposed Brain Floating Point (i.e., BF16) for training that is able to allocate more exponent bits and fewer significant bits than FP16.

\subsection{Data efficiency}
\label{sec: data efficiency}

Data efficiency represents how efficiently a training pipeline leverages its data. It determines the number of iterations (steps) required to complete a training process, thus affecting the overall training cost. Since existing LLMs such as LLaMA~\cite{touvron2023llama} are usually trained on a large quantity of texts, maximizing the utilization of data offers a promising solution to reduce training cost. 

Recent works try to improve data efficiency in various aspects of the training pipeline. We identify three major directions of achieving this goal: importance sampling, data augmentation, and training objective.

\subsubsection{Importance sampling}
\label{sec: importance sampling}
 A current survey~\cite{zhuang2023survey} notes that importance sampling (data pruning) significantly influences models’ data efficiency during pre-training. Importance sampling means to prioritize informative training instances, so it involves estimating per-sample importance. It is also called data pruning. A major solution of importance estimation is to compute gradient norm~\cite{johnson2018training,katharopoulos2018not}. More recent approaches~\cite{paul2021deep,sorscher2022beyond} work towards accelerating the data importance sampling process. 
 
 Data-Juicer~\cite{chen2023datajuicer}, an LLM data processing system, enables efficient and scalable data processing to improve the quality of the training data. As a result, the generated data recipes from Data-Juicer yielded considerable improvements on LLaMA~\cite{touvron2023llama} performance in various pre-training and post-tuning cases. Similarly, INGENIOUS~\cite{renduchintala-etal-2023-ingenious} is another system that aims to improve data quality by selecting highly representative subsets of the training corpora. ASTEROID~\cite{10.5555/3618408.3618539}, a multi-stage computational framework, first trains an MLFF (machine learning force fields) model on a large amount of inaccurate data to captures the sophisticated structures of training data and then fine-tunes the obtained model on a small amount of accurate data to improve model performance. Since inaccurate data is cheap while accurate data is much more expensive, ASTEROID improves data efficiency by fully utilizing the cheap inaccurate data. 
LISA~\cite{pan2024lisa}, a memory-efficient fine-tuning method for LLMs, leverages layerwise importance sampling to selectively update model layers, thereby reducing GPU memory consumption while outperforming traditional full-parameter tuning and LoRA in downstream tasks.

\subsubsection{Data augmentation}
\label{sec: data augmentation}
Data augmentation creates modified copies of existing data so that the current data can be fully utilized. Since it is an effective technique of improving data efficiency, a joint data augmentation for vision-language representation learning~\cite{Hao_2023_WACV} is proposed to improve the existing pre-training pipelines. As an outcome of improving the data efficiency of pre-training pipelines, researchers of this work also show that downstream performance can be positively impacted. The training of generative adversarial networks (GANs) is also benefited by data augmentation~\cite{hou2022augmentation}. Moreover, work has been done to augment acoustic data through pseudo acoustic representations of textual data to improve speech processing~\cite{lu-etal-2023-improving}. The proposed LLMRec framework enhances recommender systems by leveraging large language models to augment user-item interaction graphs, item attributes, and user profiles, thereby addressing data sparsity and improving recommendation accuracy~\cite{wei2024llmrec}. A novel data augmentation technique, LLM-DA, leverages the text generation capabilities of large language models to enhance few-shot named entity recognition by generating semantically coherent and diverse training data~\cite{ye2024llm}.

\subsubsection{Training objective}
\label{sec: training objective}
 A recent survey~\cite{kaddour2023challenges} finds that the choice of pre-training objective is another factor that determines data efficiency. For the design of pre-training objective~\cite{fan-he-2023-efficient}, it is typically a function of model architecture, input/target construction, and masking strategy. Specifically, representative masking strategies include masked language modeling~\cite{song2019mass}, masked image modeling~\cite{he2022masked}, and language-image pre-training~\cite{li2023scaling}. Researchers of these works find that skipping the processing of some masked tokens can significantly improve training efficiency.
\section{LLM fine-tuning}
\label{sec: finetune}

Fine-tuning Large Language Models (LLMs) like GPT-4 for specialized tasks involves a critical balance between achieving task-specific performance and maintaining resource efficiency, given their considerable size and computational demands. This section discusses various fine-tuning strategies, focusing on optimizing computational load, memory usage, and energy consumption. Techniques such as parameter-efficient fine-tuning, which adjusts a limited subset of parameters, offer a resource-conscious approach, while full-parameter fine-tuning, involving the modification of all parameters, is explored in the context of its higher resource requirements. This exploration is key to understanding how fine-tuning in LLMs is evolving to address the dual challenges of performance optimization and resource constraints.

\subsection{Parameter-efficient fine-tuning}
\label{sec: parameter efficient finetuning}

Parameter-efficient fine-tuning (PEFT) is a technique aimed at making the most out of an LLM's vast parameter space without the need to adjust all the parameters during the fine-tuning process. Given the immense size of modern LLMs, fine-tuning every parameter can be computationally expensive and may even risk overfitting on smaller, task-specific datasets. Current PEFT techniques can be categorized into two main streams: 1) Masking-based Fine-tuning and 2) Adapter-based Fine-tuning. 

\noindent\textbf{Masking-based fine-tuning.} In this approach, only a subset of the model’s parameters is updated during the fine-tuning process. The rest of the parameters are ``masked'' or frozen, meaning they are not updated during backpropagation. This masking could be applied to specific layers, certain types of parameters, or parameters identified through various criteria like importance scores. Previous research endeavors \cite{jiang2019smart,zaken2021bitfit} have focused on the optimization of fine-tuning procedures for comparatively smaller language models through the deployment of diverse regularization methodologies. However, these approaches exhibit limitations when applied to the fine-tuning of LLMs due to the significantly elevated computational requirements and voluminous datasets essential for the effective training of such models. In addressing these challenges, the method known as CHILD-TUNING \cite{xu2021raise} employs data from the target task to identify a subset of parameters—referred to as the ``child network''—that are most pertinent to the task, while preserving the pre-trained values for parameters in the remaining architecture. In a related vein, a list of methods \cite{zhang2022fine,yu2023unlearning, abdurrahman2023typhoon} introduces a dynamic parameter selection pipeline specifically tailored for the efficient fine-tuning of LLMs. These works adaptively designate a judicious sub-network for staged updates, leveraging gradient information from back-propagation, thereby achieving notable performance enhancements on domain-specific tasks, particularly in resource-constrained environments. To optimize resource usage, MEFT \cite{hao2024meft} implements sparse activations and a Mixture of Experts (MoE) approach, dynamically offloading trainable parameters to the CPU and selectively transferring only the most relevant ones to the GPU, thereby reducing GPU memory load and communication overhead. To tackle the rank selection challenge, DyLoRA \cite{valipour2022dylora} leverages a dynamic low-rank adaptation approach, training LoRA blocks across a spectrum of ranks and enabling rank-specific inference without the need for costly search processes, thus optimizing efficiency while maintaining performance across a range of model sizes.

\noindent\textbf{Adapter-based fine-tuning.} Different from the previous method, in this approach, additional lightweight layers (adapters) are inserted between existing layers of the pre-trained model. During fine-tuning, only the parameters of these adapter layers are updated, while the original model parameters are kept fixed \cite{houlsby2019parameter, hu2023llm}. Recent scholarly contributions have focused on Unsupervised Domain Adaptation (UDA) employing adapter mechanisms to advance the capabilities of pre-trained models in cross-lingual or multi-task learning contexts. A pioneering approach \cite{zhang2021unsupervised} involved multi-domain adaptation through a bifurcated strategy: an initial domain-fusion training phase employing Masked Language Model (MLM) loss on a composite corpus, followed by task-specific fine-tuning. A subsequent development, UDApter \cite{malik2023udapter}, extended this dual-phase methodology by compartmentalizing it into two distinct adapter modules: a domain adapter for domain-invariant feature extraction, and a task adapter with static parameters. The underlying architecture was based on AdapterFusion \cite{pfeiffer2020adapterfusion}. Another advancement, AdapterSoup \cite{chronopoulou2023adaptersoup}, further optimized the adaptation process by utilizing a weight-averaging approach on domain adapters exclusively during the evaluation stage. Various techniques for domain adapter selection were investigated, including exhaustive combination, text clustering, and semantic similarity measures.

Adapters with underlying neural network architectures are commonly referred to as \textit{neural adapters}. The seminal design of such adapters is attributed to Houlsby et al. \cite{houlsby2019parameter} and consists of a sequential arrangement of down-projection, a GeLU non-linear activation function \cite{hendrycks2016gaussian}, and up-projection. These components are integrated with feed-forward layers to serve as the foundational architecture. Subsequent work by Bapna et al.~\cite{bapna2019simple} streamlined this structure, reducing it to a single hidden-layer feed-forward network while empirically demonstrating its efficacy in domain adaptation tasks. These adapter modules are strategically positioned after the multi-head attention and feed-forward layers within the transformer architecture. Variants of such neural adapters are colloquially termed as either bottleneck adapters or serial adapters; in the present paper, we employ the term "serial adapters" to refer specifically to the architecture described in \cite{houlsby2019parameter}.

Finally, Low-rank adaptation (LoRA)~\cite{hu2021lora} is inspired by the observation that large language models reside on an intrinsic subspace~\cite{aghajanyan2020intrinsic}, where model parameters are efficiently updated. Therefore, learning in this subspace significantly reduces the amount of parameters. LoRA modules implant learnable SVD blocks as the subspace with a low matrix rank $r \ll d$, where $d$ is the dimension of input data. The matrices are added in parallel to the pre-trained weights, thus keeping them frozen during the fine-tuning. Notably, LoRA shows superiority in further reducing the number of trained parameters and introducing no latency during inference.

\subsection{Full-Parameter fine-tuning}
\label{sec: full parameter finetuning}

As the name suggests, in the paradigm of full-parameter fine-tuning, all model parameters are subject to change during training. With a higher training cost than the PEFT, full-parameter fine-tuning can generally lead to better performance than the parameter-efficient methods~\cite{sun2023comparative}. However, this phenomenon may not hold true on a simple dataset (e.g., a dataset with a lack of language diversity) for a specific downstream task~\cite{razuvayevskaya2023comparison}. Since PEFT only trains a relatively small number of parameters, models trained via full-parameter fine-tuning methods have a greater learning capacity. Moreover, the convergence of PEFT is generally not as fast as that of full-parameter fine-tuning~\cite{ding2023parameter}. As for training cost, it is reported~\cite{sun2023comparative} that ``using full-parameters fine-tuning requires about 3-5 times the time cost'' of LoRA fine-tuning. GPU memory consumption is also a concern because updating all the parameters can be impractical when dealing with LLMs. The significantly higher training cost of full-parameter fine-tuning poses a challenge for researchers in choosing which method to use. As for memory cost during training, several optimization methods have been proposed, such as Gradient Checkpointing~\cite{chen2016training}, Zero Redundancy Optimizer~\cite{rajbhandari2020zero} and Flashattention~\cite{dao2022flashattention}.

To mitigate the cost of training, many recent works on full-parameter fine-tuning aim to optimize memory consumption~\cite{lv2023parameter,malladi2023finetuning}, which significantly reduces the barrier of this research. For example, a new optimizer called LOMO (\textbf{LO}w-\textbf{M}emory \textbf{O}ptimization) was proposed~\cite{lv2023parameter} to combine gradient computation and parameter update in one training step in order to improve memory efficiency. Stochastic gradient descent (SGD) was adopted in this method, and a theoretical analysis was provided to show the effectiveness of SGD on fine-tuning all the parameters of LLMs. As a result, the full parameter fine-tuning of a 65B model requires less than 192GB GPU memory (``a single machine with 8×RTX 3090, each with 24GB memory''). LOMO presents a practical solution to train LLMs in resource-constrained scenarios. Another recently proposed optimizer called MeZO~\cite{malladi2023finetuning} estimates gradients using only two forward passes and fine-tunes LLMs ``with the same memory footprint as inference''. Requiring 55GB GPU memory, it can train a 30B model via full-parameter fine-tuning. The HiFT method~\cite{liu2024hift}, a hierarchical fine-tuning strategy, addresses GPU memory constraints in full-parameter fine-tuning of language models by updating only a subset of model parameters at each step. This block-by-block update approach enables HiFT to achieve comparable performance to conventional full-parameter fine-tuning with significant memory savings. Notably, HiFT supports the fine-tuning of 7B models on devices with 24GB memory without additional memory-saving techniques. This method, compatible with various optimizers, shows potential as a scalable and efficient solution for large language model adaptation in memory-limited environments.

Researchers are also paying attention to data-centric knowledge injection~\cite{wu2023pmcllama} when adapting a general-purpose foundation model towards a specific domain such as healthcare. The knowledge injection can be achieved by fine-tuning on domain-specific textbooks, publications, and instructions. As an identified drawback~\cite{kumar2022fine} of full-parameter fine-tuning, trained models can distort their pre-trained features and underperform on data distributions unseen during fine-tuning.
\section{LLM inference}
\label{sec: inference}

Inference in Large Language Models (LLMs) like the GPT series is a critical stage where trained models are applied to generate text, answer questions, or perform other language tasks based on their training. With the expansive size and complexity of these models, enhancing the efficiency of the inference process is essential. This section examines various techniques to optimize LLMs for inference, focusing on strategies that reduce computational load and memory usage while maintaining high-quality outputs. The approaches explored include model compression methods like pruning and quantization, and dynamic inference techniques that adaptively adjust computation based on input data. These methods are crucial for deploying LLMs in real-world applications, where resource constraints and performance requirements are key considerations.

\subsection{Model compression}
\label{sec: model compression}

Model compression and acceleration are prevalent techniques in which a cumbersome, slow-performing model is optimized to produce a streamlined version. This refined model not only requires minimal storage—making it apt for mobile device deployment—but also operates with reduced latency. Moreover, initially training a sizable model and subsequently compressing it can enhance training efficiency and bolster its generalization capabilities.

\subsubsection{Pruning}
\label{sec: pruning}

Sparsity, one of the longest-standing concepts in machine learning, was introduced to the neural network field as early as the 1980s~\cite{lecun1989optimal}. It was picked up again for “modern” deep networks in the late 2010s, first under the name of Pruning, with the primary goal of reducing inference costs~\cite{han2015deep}. In general, pruning methods can be divided into two categories, i.e., \emph{structured pruning} and \emph{unstructured pruning}. Structured pruning targets higher-granularity structures, such as entire neurons, channels, layers, or rows/columns of weight matrices. Structured pruning results in a model with reduced size that retains its original architectural structure, making it more hardware-friendly for deployment. On the other hand, unstructured pruning involves removing individual weights or connections throughout the model based on certain criteria (e.g., smallest magnitude weights). Unstructured pruning produces a model with "holes" or sparse weight matrices, which require specialized software or hardware for efficient deployment. Recent research efforts have been devoted to combining LLMs with pruning techniques, aiming to tackle the substantial size and computational costs associated with LLMs.

\noindent\textbf{Unstructured pruning.}
Unstructured pruning reduces the complexity of an LLM by eliminating particular parameters without taking into account its intrinsic organization. This method focuses on individual weights or neurons in the LLM, typically by setting a threshold and nullifying parameters beneath it. Yet, by not respecting the overarching structure of the LLM, it leads to a model with a non-uniform sparse makeup. This non-uniformity necessitates unique compression methods to effectively store and compute the trimmed model. 
SparseGPT~\cite{frantar2023massive} represents a rapid unstructured pruning technique designed specifically for LLMs with hundreds of billions of parameters, allowing for operation within mere hours. Remarkably, it can reduce parameters by as much as 60$\%$ without compromising the model's performance significantly. 
To address the demanding weight update process of SparseGPT, Wanda~\cite{sun2023simple} introduces a novel pruning criterion. Wanda assesses each weight by calculating the product of its magnitude and the norm of its related input activations, using an estimation from a concise calibration dataset. This criterion is used for intra-layer comparisons in linear layer outputs, facilitating the exclusion of less significant weights in LLMs.
In \cite{tuli2023acceltran}, the author proposes a novel dynamic inference scheme, DynaTran, which prunes activations at runtime with low overhead, which improves the throughput of transformer inference. The author designed an application-specific integrated circuit (ASIC) based architecture called AccelTran to implement the DynaTran. Specifically, several hardware-aware designs are proposed including matrix tiling and various dataflow to improve data reuse.
Bai et al proposed SparseLLM~\cite{bai2024gradient}, a novel global pruning framework for LLMs. Unlike prior methods limited to layer-wise sparsity, SparseLLM achieves global pruning with a novel optimization design and demonstrates that global pruning can retain model accuracy while reducing resource demands, making it advantageous for deployment in large-scale, resource-constrained environments.

\noindent\textbf{Structured pruning.} 
Structured pruning involves the selective removal of an entire group of weights. The definition of ‘group’, which makes those amenable to hardware speedup, could refer to weight blocks, neurons, filters/channels, attention heads, or other dedicated fine-grained sparse patterns. 
\cite{ma2023llm} introduce LLM-Pruner, a pioneering framework tailored for structured pruning of LLMs offering task-agnostic compression and efficient data usage. LLM-Pruner integrates a dependency detection mechanism to identify interconnected structures in the model. It utilizes an effective importance estimation approach, combining both first-order data and estimated Hessian information. This approach streamlines the selection of prime groups for pruning, enhancing the compression procedure. 
\cite{li2023losparse} propose LoSparse (Low-Rank and Sparse approximation), a novel model compression technique that approximates a weight matrix by the sum of a low-rank matrix and a sparse matrix. Pruning enhances the diversity of low-rank approximations, and low-rank approximation prevents pruning from losing too many expressive neurons. 
\cite{tao2023structured} further considers pruning the hidden dimension (e.g., embedding layers, layer normalization) of LLM besides pruning the attention heads and feed-forward layers. 
\cite{kurtic2023ziplm} proposed a new structured compression approach for LLMs, called ZipLM, which provides state-of-the-art compression-vs-accuracy results,
while guaranteeing to match a set of (achievable) target speedups on any given target hardware. Specifically, given a task, a model, an inference environment, as well as a set of speedup targets,
ZipLM identifies and removes redundancies in the model through iterative structured shrinking of
the model’s weight matrices.

\noindent\textbf{Contextual pruning}
While sparsity stands as a viable strategy to mitigate the burden in LLM inference, existing techniques either necessitate expensive retraining, compromise the LLM's intrinsic learning capabilities, or fail to accelerate real-time performance on contemporary hardware.
Zichang Liu et al.~\cite{liu2023deja} postulate that the application of \emph{contextual sparsity} — utilizing small, input-dependent sets of attention heads and MLP parameters to approximate the dense model's output — can overcome these challenges. Their investigations confirm the presence of contextual sparsity and its potential for precise prediction, enabling us to leverage it to hasten LLM inference without sacrificing model quality or learning abilities in context.
To capitalize on these findings, they introduce Deja Vu, a system proficient in dynamically predicting contextual sparsity using a cost-effective algorithm. This system, coupled with an asynchronous and hardware-optimized execution, significantly accelerates LLM inference times. 

% Experimental validations affirm that Deja Vu markedly diminishes the inference latency of OPT-175B, achieving over twice the speed of the leading FasterTransformer and surpassing the prevalent Hugging Face implementation by over six times, all without undermining model efficacy.

\subsubsection{Quantization}
\label{sec: quantization}

The quantization-based approach aims to achieve substantial model compression at the cost of affordable loss of model accuracy. While the conventional method for representation learning adopts floating-point numbers, quantization converts them to fewer bits such as integers or other discrete numbers, making models more efficient regarding both memory and computation, especially suitable for deployment on resource-constrained devices. Although this might lead to the loss of model precision (or quantization error) to some extent, careful quantization techniques can achieve substantial model compression with only minimal accuracy degradation. Based on which module of the model the quantization is applied to, quantization-based methods can be classified into four scenarios: (1) weight quantization, (2) activation quantization, and (3) fixed-point quantization. 

% which will be summarized in Table~\ref{tab: quantization table} and introduced in detail as follows.

\noindent\textbf{Weight quantization.}
The most popular practice for quantization is weight quantization, which compresses language models by representing model weights using fewer bits. For example, Lee et al.~\citep{lee2023flexround} jointly learned a common quantization grid size and the division factor for pre-trained weights and performed element-wise division on weights. Instead of quantizing all weights of the model which may lead to moderate-to-high quantization error, another natural thought is to identify and quantize weights that are not important. Some works perform weight quantization after the training process. For instance, Lin et al.~\citep{lin2023awq} identified and preserved only 1$\%$ of salient weights by observing activation that can largely reduce the quantization error. Dettmers et al.~\citep{dettmers2023spqr} and Wei et al.~\citep{wei2023outlier} identified and isolated outlier weights that potentially lead to large quantization error via different techniques such as filtering sensitivity-based algorithm~\citep{dettmers2023spqr} or identifying asymmetric presentation and scaling down problematic channels~\citep{wei2023outlier}. Some work leverages either activation or model outliers sensitive to accuracy degrade from weight quantization. For instance, Kim et al.~\citep{kim2023squeezellm} also employed a sensitivity-based that searches for optimal bit precision assignment and stores outliers and sensitive weight values in an efficient sparse format. Lee et al.~\citep{lee2023owq} studied how activation outliers can amplify the error in weight quantization and assign higher precision to the weights susceptible to quantization caused by activation outliers. Guo et al.~\cite{guo2023olive} handled outlier values locally by sacrificing values next to outliers (usually not important) to accommodate those important outliers. Liu et al.~\cite{liu2023llm} focused on the weight quantization of generative models by applying distillation based on generations produced by the pre-trained model. Frantar et al.~\citep{frantar2022gptq} proposed a one-shot weight quantization method based on approximated second-order information, reducing the bandwidth of the GPT model down to 3 or 4 bits per weight. Lin et al.~\citep{lin2024rotation} proposed DuQuant, a novel quantization strategy that employs rotation and permutation transformations to manage activation outliers more effectively, achieving state-of-the-art performance for low-bit weight-activation quantization on various large language models (LLMs). Shao et al.~\citep{shao2023omniquant} proposed OmniQuant, a quantization method leveraging learnable weight clipping and equivalent transformations, effectively optimizing quantization for large language models while achieving state-of-the-art performance across various low-bit quantization settings. Some works achieve weight quantization during training. For example, Yang et al.~\citep{yang2023dynamic} proposed dynamic stashing quantization that dynamically quantizes the intermediate results between forward and backward processes for a significant reduction of the memory traffic during training. Yang et al.~\citep{yang2023quantization} used low-rank tensor train and tensor-train matrix formats to represent the embedding tables and linear layers during training. Dettmers et al.~\citep{dettmers2023qlora} backpropagated gradients through a frozen, 4-bit quantized pre-trained language model into Low-Rank Adapters (LoRA) and performed double quantization by quantizing quantization constants. Wortsman et al.~\citep{wortsman2023stable} accelerated and stabilized large language-vision models by reducing the weights to low-bit values, such as using 16-bit precision for weight gradient computation and int8 multiplications for the forward pass and layer input gradient computations. Other works approach to focus on the pre-trained model. Gong et al.~\citep{gong2023prequant} quantized the pre-trained model in a task-agnostic way to obtain a ``pre-quantized” model before fine-tuning and froze most of the quantized weights in the “pre-quantized” model.

\noindent\textbf{Activation quantization.}
In addition to weight quantization, other techniques such as activation quantization and fixed-point quantization have been employed to ease the heavy memory consumption handling LLMs. Activation quantization deals with quantizing the intermediate values (i.e., activations) that arise during model inference. For instance, Liu et al.~\citep{liu2022gact} proposed a framework agnostic to the neural work architecture by approximating the gradient descent of activation compression training~\citep{evans2021ac} via a linearized
version. Liu et al.~\citep{liu2023llm} not only performed weight quantization but also quantized activations to 6-bit precision. 

\noindent\textbf{Fixed-point quantization.}
Fixed-point quantization represents weights and activations using fixed-point arithmetic to reduce memory usage and accelerate computations. Yu et al.~\citep{yu2023boost} pruned transformer-based language models to meet the GPU’s acceleration constraint of structured sparse patterns with FP16 type. Then the floating-point sparse model is quantized into a fixed-point one by quantization-aware training.

\subsubsection{Knowledge distillation}
\label{sec: knowledge distillation}

The distillation of domain-specific knowledge from LLMs into more compact neural networks has emerged as a promising area. Considering the specialty of LLMs, recent techniques of Knowledge Distillation can be divided into two streams: 1) \textit{White-box Knowledge Distillation}: the teacher model's parameters are available to
use; 2) \textit{Black-box Knowledge Distillation}: only the
teacher model's predictions are accessible.

\noindent\textbf{White-box knowledge distillation.}
This approach not only aims to substantially decrease inference latency but also to amplify the effectiveness of specialized task-solving capabilities. A compelling example is the work by Muhamed et al., who ingeniously compressed a behemoth 1.5 billion-parameter white-box LLM into a far more manageable 70 million-parameter model. This was specifically engineered for optimizing Click-Through Rate (CTR) prediction tasks. They introduced an innovative architecture featuring twin-structured BERT-like encoders coupled with a fusion layer. This allowed for a seamless cross-architecture knowledge distillation from a single LLM, yielding superior performance metrics in both real-time online and controlled offline environments \cite{muhamed2021ctr}. In a parallel vein, several studies \cite{vucetic2022efficient,marjieh2023language,azerbayev2022explicit,gu2023knowledge,zhang2023not} have also made strides in this field by incorporating a specialized knowledge distillation module during the fine-tuning process of LLMs. This results in a twofold benefit: accelerated convergence rates and more efficient utilization of computational resources. The distillation module intelligently leverages pre-trained model parameters to expedite the convergence process, while concurrently training a selective subset of parameters to effectively counteract the issues associated with model over-parameterization. Extending this concept further, additional works \cite{shridhar2022distilling,hsieh2023distilling} have ventured into the intricate process of distilling the nuanced chain-of-thought reasoning capabilities inherent in larger models into their smaller counterparts. This allows the miniaturized models to inherit a form of \textit{cognitive reasoning} from their more oversized progenitors, thereby enhancing their overall utility and performance.

\noindent\textbf{Black-box knowledge distillation.} 
Another line of research on Knowledge Distillation focuses on the somewhat elusive task of distilling knowledge from ``black-box'' large language models (LLMs) like ChatGPT. In these cases, researchers are often limited to interacting only with the model's predictions, without the luxury of directly accessing its internal parameters or architecture. This is a particularly challenging endeavor because the traditional methods of knowledge distillation, which often rely on structural similarities or parameter sharing between the teacher and student models, are rendered inapplicable. These types of works \cite{wu2023lamini,peng2023instruction,chiang2023vicuna,taori2023stanford,ling2023open} have leveraged LLMs as a query generation machine that directly generate \textit{high quality} instruction following queries (and answers) to fine-tune smaller LLMs (e.g., LLaMA). The obtained smaller LLMs exhibit a stronger instruction-following capability.

\subsubsection{Low-rank approximation}
\label{sec: low rank approximation}

Due to low memory cost, low-rank approximation has made the model compression more viable and practical. A common approach is singular value decomposition (SVD). For a low-rank matrix $A \in R^{m\times n}$, where $r$ is the rank of matrix $A$, there exists $U \in \mathbb{R}^{m\times r}$, $V \in \mathbb{R}^{n\times r}$ are two orthogonal matrices; $\sigma \in \mathbb{R}^{r\times r}$ is a diagonal matrix with only the non-zero singular values of $A$. Through SVD, we reduce the memory cost from $O(mn)$ to $O((m+n)\times r)$, which is a huge saving in many scenarios.

In general, any linear matrix can be approximated through SVD. 
\cite{wang2023lighttoken} omit the diagonal matrix in SVD decomposition and encode the residue of the original matrix and approximated matrix to achieve better performance. 
\cite{zhang2023pruning} used low-rank matrices to evaluate the parameter importance. They utilize low-rank matrices to formulate the optimization problem and solve it to get the approximation of the original parameter. \cite{wu2023zeroquant} applied low-rank approximation to reduce quantization errors. They use low-rank decomposition to reduce error without a huge impact on the speed of inference of LLM. \cite{chen2021drone} achieved low-rank approximation through the observation that data of NLP task is always in low-rank subspace. They first decompose the matrix of Feed-forward propagation through SVD and solve the optimization problem to get the needed low-rank matrices. \cite{tahaei2021kroneckerbert} and  \cite{edalati2021kronecker} use conduct decomposition for layers in the transformer and GPT-2 respectively through the Kronecker product. It is a new way of "multiplication" different from the traditional matrix multiplication. \cite{reid2021subformer} utilize low-rank approximation to reduce the parameters of generative transformers up to 25\%. The non-contextual embeddings will have far fewer
 features compared with contextual ones, which is a huge saving for large language models. \cite{xu2023tensorgpt} tackles the storage problem of large language models through low-rank approaches. They store embeddings in low-rank format to reduce the memory cost, making the deployment of LLM in edge devices possible. \cite{gao2024dlora} introduces DLoRA, a distributed fine-tuning framework for large language models that enhances parameter efficiency and privacy. By offloading fine-tuning tasks between cloud and edge devices, DLoRA addresses the limitations of purely cloud or edge-based solutions, ensuring data privacy and reducing computation and communication costs. The Kill and Revive algorithm further optimizes performance by dynamically tuning only the most responsive parameters, achieving significant reductions in workload while maintaining accuracy on downstream tasks. \cite{lin2024splitlora} presents SplitLoRA, an efficient fine-tuning framework that combines split learning with federated learning to address large model training burdens. By partitioning the model, SplitLoRA reduces computational demands on client devices while maintaining model accuracy. This framework, which uses LoRA for parameter-efficient tuning, achieves faster convergence and lower resource use compared to traditional federated approaches, making it suitable for deployment in resource-limited environments. \cite{lin2024data} introduces DEALRec, a data-efficient fine-tuning method for LLM-based recommendation systems. This approach optimizes few-shot fine-tuning by selecting influential samples that are representative of full data, balancing both influence and effort scores to maximize accuracy with minimal data. DEALRec, tested on three real-world datasets, achieved superior performance over full-data fine-tuning while significantly reducing computational costs, making it effective for dynamic recommendation environments. ~\cite{li2024evaluatingquantizedlargelanguage} explores post-training quantization (PTQ) as a solution for reducing memory and computational demands of large language models (LLMs). PTQ is applied across three types of tensors—Weights, Activations, and KV Cache—to optimize efficiency while assessing impact on model performance. The study evaluates models from 11 families, such as LLaMA2, Falcon, and Vicuna, across five task categories, including basic NLP, dialogue, and long-context tasks. Key findings suggest specific bit-width quantization strategies that balance performance and efficiency, revealing trends in performance degradation across tensor types and model sizes. This comprehensive evaluation serves as a guide for selecting quantization methods suited to different LLM applications and offers insights into optimizing model deployment under resource constraints. 
 ~\cite{hu2024illmefficientintegeronlyinference} introduces a new post-training quantization (PTQ) method specifically designed for LLMs to run with integer-only operations, aiming to eliminate floating-point computations. Key components include Fully-Smooth Block-Reconstruction (FSBR) to stabilize inter-channel variations, Dynamic Integer-only MatMul (DI-MatMul) for dynamic quantization in matrix multiplication, and specialized integer-only non-linear operators like DI-ClippedSoftmax. This framework achieves significant inference efficiency while retaining accuracy comparable to floating-point models, demonstrating I-LLM's potential for resource-limited deployments on edge devices. ~\cite{zeng2024abqllmarbitrarybitquantizedinference} introduces an innovative quantization framework designed to enable high-performance inference for LLMs under various bit-precision configurations. ABQ-LLM tackles challenges in quantized inference, such as performance degradation at low bit widths and limited support for non-standard precision formats on GPUs. Key innovations include distribution correction methods to handle quantization-induced distribution shifts and a bit balance strategy to reduce asymmetry issues in low-bit quantization (e.g., INT2). ABQ-LLM outperforms existing methods like SmoothQuant and I-LLM, demonstrating significant acceleration and memory efficiency, especially in configurations like W2A8 for LLaMA models.

\subsection{Dynamic acceleration}
\label{sec: dynamic inference}

In Section~\ref{sec: model compression}, we have introduced techniques for reducing the number of parameters in an LLM for inference acceleration. These methods are general and agnostic to input data, i.e., \emph{static} for any given input sequence. However, there is another line of methods that aims to improve the efficiency of LLM inference without reducing the number of parameters. Such methods typically are specific to different input sequences and we term them as \emph{dynamic acceleration} methods. In general, existing dynamic acceleration methods include 3 categories, i.e., \emph{early exit}, \emph{token pruning}, and \emph{token parallelism}. Early exist accelerates model inference by terminating inference at a particular layer-based
on some criteria, i.e., making an LLM shallower. On the other hand, token pruning accelerates inference by skipping some tokens for higher layers based on their importance, i.e., making an LLM input shorter. Last, token parallelism considers leveraging certain techniques or algorithms to generate multiple tokens in parallel (opposite to autoregressive fashion that generates each token sequentially).

\subsubsection{Early exit}
\label{sec: early exit}
Early exit is an inference acceleration strategy used in neural networks by skipping the computation of certain layers. The rationale behind early exit is that simper input samples usually require less calculation to make predictions~\cite{teerapittayanon2016branchynet,jawahar2019does,rogers2021primer}.
Pioneering explorations on early exit often rely on defining their own early-exit criterion: DeeBERT~\cite{xin2020deebert} adapts entropy as its exit criterion; RightTool~\cite{schwartz2020right} adapts softmax scores of prediction as its exit criterion; PABEE~\cite{zhou2020PABEE} exit inference when the intermediate predictions of the internal classifiers remain unchanged consecutively.
PCEE-BERT~\cite{zhang2022pcee} proposes a hybrid early exit criterion that combines confident score with patience counter. In other words, PCEE-BERT will early exit when enough numbers of consecutive intermediate layers are confident.
SkipBERT~\cite{wang2022skipbert} accelerates inference by skipping the computation of shallow layers when precomputed text chunks are met. The Higher layers can be further skipped using the early-exit criterion.
Short-Cutting Transformer~\cite{din2023jump} suggests a linear transformation-based method to cast intermediate representations as final representations, thus bypassing the transformer computation in between. Short-Cutting Transformer adapts the same early exit strategy as in CALM~\cite{schuster2022confident}, where the LLM early exits when the difference between the highest and the second highest probabilities is bigger than CALM's confidence threshold.
MuE~\cite{tang2023you} extends dynamic early exit strategy to multimodal LLMs. Unique challenges arise since existing early exit strategies can not directly apply to the widely-used unified multimodal architecture with both encoder and decoder, due to the difficulty of making exit decisions when dependencies between encoder and decoder exit. MuE proposes its exit criterion based on the layer-wise input similarity, inspired by the saturation observation~\cite{geva2022transformer}.

\subsubsection{Input pruning}
\label{sec: token pruning}

Input Pruning explores the opportunity for the dynamic reduction of input sequence length to improve the Transformer’s computational efficiency. Its intuition is similar to the human being’s reading comprehension capability it does not read all words equally. Instead, some words are focused with more interest while others are skimmed. For Transformer models, this means adopting a dynamic computation budget for different input tokens according to their contents.

Existing input pruning works can be categorized into two classes based on token removal or retention criteria. The first class uses value-based scoring (e.g., attention) to identify unimportant tokens. For instance, SpAtten~\cite{wang2021spatten} ranks tokens using importance scores and retains the top-k highest-scoring tokens. LTP~\cite{kim2022learned} improves PoWER-BERT by introducing a learnable layer-wise threshold, enabling adaptive pruning length. ToP~\citep{li2023constraint} overcomes the limitation of inaccurate token importance ranking in the self-attention mechanism through a ranking-distilled token distillation technique, which distills effective token rankings from the final layer of unpruned models to early layers of pruned models.

The second class of token pruning methods inserts a prediction module before each transformer layer to provide a more accurate token importance score prediction. TR-BERT~\citep{ye2021tr} introduces a dynamic mechanism for making decisions about skipping tokens. It is trained with reinforcement learning with a reward that promotes classifier confidence and penalizes the number of retained tokens. Transkimmer~\citep{guan2022transkimmer} is a notable example that inserts a 2-layer MLP network at each layer as the prediction module. However, the extra prediction module can also introduce considerable inference latency overhead, which is unfriendly on resource-limited devices. PuMer~\citep{cao2023pumer} proposed a token reduction framework that uses text-informed pruning and modality-aware merging strategies to progressively reduce the tokens of input image and text, improving model inference speed and reducing memory footprint. PuMer learns to keep salient image tokens related to the input text and merges similar textual and visual tokens by adding lightweight token reducer modules at several cross-modal layers in the Vision-Language model. Infor-Coef~\citep{tan2023infor} proposes a model acceleration approach for large language models that incorporates dynamic token downsampling and static pruning, optimized by the information bottleneck loss. The token sampler, which is similar to the MLP module of Transkimmer, is trained for downsampling the token length before the multi-head attention layer. SMART-TRIM~\cite{wang2023smarttrim} incorporates lightweight trimming modules (MLP layers) into the backbone to perform task-specific pruning on redundant inputs and parameters, without the need for additional pre-training or data augmentation. LLMLingua-2~\citep{pan2024llmlingua} proposes a task-agnostic prompt compression method by formulating it as a token classification task. Despite its small size, it achieves notable speedups, reducing latency by 1.6x-2.9x with a compression ratio of 2x-5x, while preserving crucial information for effective prompt understanding across various downstream tasks. Compressed Context Memory (CCM)~\citep{kim2023compressed} introduces a dynamic context compression mechanism for language model inference by integrating lightweight LoRA during forward passes, achieving a memory-efficient solution for expanding context without fine-tuning the entire model. However, CCM-concat’s higher memory demands at later time steps may be challenging for memory-constrained environments. GRIFFIN~\cite{dong2024promptpromptedadaptivestructuredpruning}, introduced in Prompt-prompted Adaptive Structured Pruning for Efficient LLM Generation, is a training-free and calibration-free pruning method targeting transformer feedforward blocks for faster, memory-efficient LLM inference. Exploiting the phenomenon of “flocking,” where neurons show similar activations across tokens in a sequence, GRIFFIN selects key neurons during the prompt phase, maintaining model performance even with 50\% parameter reduction. Compared to magnitude pruning and MoEs, GRIFFIN achieves comparable efficiency without training overhead, showing 1.25x-1.29x speed-ups on Llama 2 and Gemma models across classification and generation tasks. LazyLLM~\cite{fu2024lazyllmdynamictokenpruning} is a dynamic token pruning method aimed at improving LLM inference efficiency for long contexts. Unlike static pruning, LazyLLM selectively calculates key-value (KV) pairs only for tokens essential to predicting the next token at each generation step. By progressively pruning tokens during both the prefilling and decoding stages, LazyLLM reduces time-to-first-token (TTFT) and overall generation time without sacrificing model accuracy. Tests on the Llama 2 model show a 2.34× TTFT speedup on multi-document QA, validating LazyLLM’s capability to accelerate LLM inference efficiently without fine-tuning

\subsubsection{Token parallelism}
\label{sec: token parallel}

Inference from large autoregressive models like Transformers is slow - decoding K tokens takes K serial runs of the model. Recent works proposed to leverage techniques such as \emph{speculative execution}~\cite{burton1985speculative} to achieve parallel generation of multiple tokens instead of a sequential manner. Leviathan et al.~\cite{leviathan2023fast} introduces ``speculative decoding," an algorithm that accelerates the sampling process from autoregressive models like Transformers by computing several tokens in parallel without altering the output distribution. This is achieved by utilizing approximation models (smaller than the original LLM) to generate speculative prefixes, which are then expanded by the larger target model, thereby accelerating the inference process without compromising the output quality. SpS~\cite{chen2023accelerating} follows a similar idea and proposed speculative sampling, which generates multiple tokens per transformer call and uses a modified rejection sampling method. SpS maintains the output distribution while accelerating the process by 2 to 2.5 times without altering the model itself. Spector et al.~\cite{spector2023accelerating}  introduces an enhanced algorithm called staged speculative decoding to expedite inference in large language models (LLMs), particularly in small-batch, on-device scenarios. The method consists of structuring the speculative batch as a tree to decrease generation costs and employing an additional stage of speculative decoding to boost performance. These improvements collectively yield a 3.16x reduction in single-batch decoding latency for a 762M parameter GPT-2-L model without compromising output quality.

% \section{Hardware Optimization}

\section{System design}
\label{sec: system design}

System design is critical in optimizing Large Language Models (LLMs) like the GPT series for efficient inference, particularly in resource-constrained environments. This section explores key strategies such as hardware offloading, which manages computational resources by leveraging different storage hierarchies, and collaborative inference, which pools resources for enhanced processing capabilities. It also examines the adaptation of LLMs for edge devices, highlighting the importance of system design in maximizing the efficiency and scalability of LLMs across various deployment scenarios.

\subsection{Deployment optimization}
\label{sec: deployment optimization}

\noindent\textbf{Hardware offloading.}
Hardware offloading means transferring temporarily unneeded data in LLM from faster accelerators to slower yet ample primary and secondary storage, such as CPU memory and disk. 
These data are subsequently reloaded as needed. This method allows large LLMs to operate efficiently on GPUs with restricted memory capacity. However, the offloading and reloading processes inherently introduce significant communication overhead. Therefore, effective offloading strategy plays a crucial role in enhancing overall system efficiency. FlexGen~\cite{sheng2023flexgen} can achieve high throughput by developing a linear programming-based search algorithm that can identify the optimal offloading strategy within a defined search space of possible offloading strategies. FlexGen further improves throughput by compressing both the weights and KV cache for LLMs down to 4 bits. Additionally, FlexGen can be extended to a multi-GPU setting by adopting pipeline parallelism. DeepSpeed~\cite{aminabadi2022deepspeed} introduced an innovative technique called ZeRO-Inference to minimize latency and enhance throughput. This approach involves pining the model weights in either DRAM or NVMe and dynamically streaming each layer into GPU memory for computation as required. For a multi-GPUs environment, DeepSpeed leverages both tensor parallelism and pipeline parallelism to attain optimal performance. FastServe~\cite{wu2023fast} designs the offloading strategy based on the priority of the jobs. Key-value tensors associated with lower-priority jobs are offloaded to host memory, while key-value tensors needed for imminent use are loaded in advance. 
LUT Tensor Core~\cite{mo2024lut} introduces a novel software-hardware co-design to accelerate low-bit LLM inference through a LUT-based mixed-precision General Matrix Multiplication (mpGEMM) approach. This design leverages operator fusion to minimize precompute overhead, table symmetrization to reduce memory storage requirements, and a bit-serial architecture for flexible precision combinations, ultimately achieving superior compute density and energy efficiency compared to conventional Tensor Cores.
BrainTransformers~\cite{tang2024braintransformerssnnllm} implements Transformer-based LLMs using Spiking Neural Networks (SNN), introducing SNN-compatible components and synaptic plasticity mechanisms, achieving competitive performance on benchmarks like MMLU (63.2) and GSM8K (76.3) while offering potential energy efficiency and biological plausibility.
Ripple~\cite{wang2024rippleacceleratingllminference} optimizes LLM inference on smartphones by leveraging neuron co-activation to reorganize neuron placement in flash memory, achieving up to 5.93× reductions in I/O latency through a two-stage solution combining offline co-activation pattern analysis and online caching strategies, addressing the intersection of sparsity-driven algorithms and storage-level system co-design.
TorchTitan~\cite{liang2024torchtitanonestoppytorchnative} introduces a PyTorch-native distributed training system with modular 3D parallelism, elastic scaling, and hardware-software co-optimization, achieving up to 30% additional acceleration for training Llama 3.1 405B on 512 NVIDIA H100 GPUs by integrating techniques like Float8 training and SymmetricMemory.

\noindent\textbf{Collaborative inference.}
Collaborative inference involves the cooperative effort of multiple users or systems working collectively to conduct inference tasks for LLMs. Each participant contributes their resources, such as computing power or data. This collaborative approach serves to mitigate the constraints of individual users or systems, ultimately leading to more efficient and accurate inference when dealing with LLMs. PETALS~\cite{borzunov2022petals} is a system that facilitates collaborative inference and fine-tuning of LLMs through online collaboration among multiple users. Each participant in this system can take on the roles of a server, a client, or both. Servers host specific model layers and respond to client requests. Also, an 8-bit compression technology is used for lowering the computational resources requirement and improving efficiency.

\subsection{Support infrastructure}

\noindent\textbf{Libraries.}
In this section, we introduce several famous frameworks for LLMs. Microsoft introduced DeepSpeed~\cite{aminabadi2022deepspeed} as a cutting-edge deep learning optimization library that incorporates a range of innovative technologies, including ZeRO and 3D Parallelism. These technologies enable efficient and effective training, inference, and compression tasks. Megatron-LM~\cite{shoeybi2019megatron}, developed by NVIDIA, is another famous framework for LLMs. It implements a combination of model and data parallelism and focuses on multi-node pre-training of transformer-based models like GPT. DeepSpeed offers an enhanced version of Megatron-LM that includes additional features such as MoE model training and Curriculum Learning. Like DeepSpeed and Megatron-LM, Colossal-AI~\cite{li2023colossal} provides multi-level parallelism strategies for large-scale distributed training. These strategies encompass Auto-Parallelism, data, tensor, pipeline, and sequence parallelism. Additionally, its heterogeneous memory management component enhances training efficiency in distributed environments with heterogeneous devices. Mesh-TensorFlow~\cite{shazeer2018mesh} is a user-friendly framework seamlessly integrated with TensorFlow, specializing in model parallelism with a primary emphasis on distributed tensor computations. GPT-NeoX~\cite{gpt-neox-library} builds upon Megatron-LM and incorporates innovations from DeepSpeed. MaxText\footnote{https://github.com/google/maxtext} is an open source open-source framework that is designed for Google Cloud TPUs. Alpa~\cite{zheng2022alpa} can automatically parallelize users' single-device code on distributed clusters with data, operator, and pipeline parallelism. Transforming the Hybrid Cloud for Emerging AI Workloads~\cite{chen2024transforminghybridcloudemerging} envisions a full-stack co-design of hybrid cloud systems incorporating generative and agentic AI, edge-to-cloud integration, quantum acceleration, and unified abstractions, aiming to create secure, efficient, and sustainable platforms that foster breakthroughs in AI-driven research and applications across academia, industry, and society.LLMServingSim~\cite{Cho_2024} simulates LLM inference serving at the granularity of iterations to capture dynamic workload variations, leverages computation redundancies across decoder blocks to avoid repetitive simulations, and provides a flexible framework for exploring heterogeneous processor designs, achieving less than 14.7% error and 91.5x faster simulation speeds than existing accelerator simulators.

\noindent\textbf{Edge devices.}
In recent research, there is a growing interest in deploying LLMs on edge devices, which typically have limited computational resources. Xu et al.~\cite{xu2023training} explore the use of techniques like Low-Rank Adaptation and Noise Contrastive Estimation to reduce the memory demands of LLMs when running them on edge devices. Furthermore, they introduce a novel approach for reducing noise by minimizing data payload size within the setting of Differential Privacy and Federated Learning (FL). Woisetschl{\"a}ger et al.~\cite{woisetschlager2023federated} gives a comprehensive analysis of the feasibility of conducting Federated Learning with LLMs on contemporary edge computing systems, offering a comparison to traditional data-centralized computing approaches. Shen et al.~\cite{shen2023large} leverages cloud-deployed LLMs to coordinate models that meet user needs and perform training via edge federated learning. EdgeFormer~\cite{ge2022edgeformer} interleave attention modules with a shared feed-forward network in the decoder layer to achieve cost-effective parameterization. ProFormer~\cite{sankar2021proformer} utilizes the LSH projection layer to replace traditional embedding lookup tables, thereby mitigating the need for extensive memory resources. GhostBERT~\cite{huang2021ghostbert} employs 1-dimensional convolutions to generate additional features, thereby mitigating memory and computational expenses. SqueezeBERT~\cite{iandola2020squeezebert} adopts optimization techniques used in computer vision networks, such as grouped convolutions, to achieve notable speedups. LiteTransformer~\cite{wu2020lite} reduces the computation of the transformer base model by isolating local feature extraction from global feature extraction. MobileLLM~\cite{liu2024mobilellmoptimizingsubbillionparameter} achieves significant accuracy improvements for sub-billion parameter models by prioritizing depth over width and leveraging techniques like weight-sharing and grouped-query attention to enhance performance with minimal memory overhead. 
EdgeShard~\cite{zhang2024edgeshardefficientllminference} optimizes LLM inference by partitioning models across heterogeneous edge devices and cloud servers, leveraging collaborative edge computing to reduce latency and increase throughput. Any-Precision LLM~\cite{park2024anyprecisionllmlowcostdeployment} leverages post-training quantization and incremental upscaling to minimize deployment costs by supporting multiple quantized LLMs of varying bit-widths, all within a memory footprint comparable to a single model. The Breakthrough Memory Solutions~\cite{10477465} enhance LLM performance by integrating processing-in-memory (PIM) and processing-near-memory (PNM) technologies, significantly boosting memory bandwidth and reducing energy consumption for large-scale AI applications. MELTing point~\cite{laskaridis2024meltingpointmobileevaluation} enables efficient mobile execution of LLMs by leveraging a benchmarking infrastructure that traces memory and energy usage, optimizing for on-device inference across various mobile and edge devices. The proposed model in ~\cite{xu2022traininglargevocabularyneurallanguage} reduces the communication overhead and memory consumption in federated learning by introducing partial embedding updates and low-rank adaptation, enabling efficient training of large-vocabulary models on resource-constrained devices. 
LLMS~\cite{yin2024llmservicemobiledevices} reduces context switching latency in mobile LLM services by employing chunk-wise KV cache compression, recompute pipelining, and lifecycle management, enabling efficient large language model execution on resource-constrained devices. 
LocMoE~\cite{li2024locmoelowoverheadmoelarge} optimizes large language model training by introducing locality-based routing and communication strategies, reducing training time and enhancing load balance without sacrificing model accuracy.
JetMoE~\cite{shen2024jetmoereachingllama2performance} reduces inference computation by leveraging sparse activation in both attention and feed-forward layers, activating only a subset of experts for each input token, leading to a 70\% reduction in computation compared to dense models like Llama2-7B.
Bai et al proposed FedSpaLLM~\cite{bai2024fedspallm}, an innovative approach to pruning large language models in federated learning settings, addressing the unique challenges of model and system heterogeneity. Key contributions include an aggregation function based on the $\ell_0$-norm for managing diverse pruning masks across clients and a novel layer-sampling strategy that ensures efficient resource utilization while maintaining the accuracy of the global model. This framework allows for a flexible yet robust pruning process, enhancing model personalization and scalability in federated environments.
FusionLLM~\cite{tang2024fusionllmdecentralizedllmtraining} enables efficient decentralized training of large DNNs across geo-distributed GPUs by representing models as operator DAGs, using an OP-Fence scheduler, and implementing adaptive compression with AdaTopK, achieving up to 9.39x speedup in training ResNet-101 and GPT-2 on heterogeneous networked environments.
Chat AI~\cite{doosthosseini2024chataiseamlessslurmnative} seamlessly integrates HPC infrastructure and cloud-based web services for privacy-preserving, real-time LLM serving, leveraging Slurm’s batch scheduling to efficiently utilize GPU resources and incorporating an SSH ForceCommand-based circuit breaker for enhanced security.

\subsection{Other systems}

Tabi~\cite{wang2023tabi} proposes an inference system with a multi-level inference engine to reduce the inference latency of LLMs. Instead of using the same model for all the queries, Tabi avoids invoking costly LLMs by employing multiple DNNs to handle heterogeneous queries within a task. Z. Peng et al.~\cite{peng2023near} leverage leveraging min-hash technique to improve the efficiency and scalability of near-duplicate sequence search for LLMs. 

\section{Technique categorization by resources}
\label{sec: application}

\begin{sidewaystable}[htbp]
\centering
\begin{tabular}{ccccccc}
\hline
Main Category Technique & Sub-Category & Computation & Memory & Energy & Money & Communication \\ 
\hline

\multirow{2}{*}{Transformer Architecture~\S\ref{sec: efficient transformer architecture}} 
& Approximated Attention & \pmb{\checkmark} & \pmb{\checkmark} & \checkmark & \checkmark & \\ 
\cline{2-7}
& Hardware-aware Attention & \pmb{\checkmark} & \pmb{\checkmark} & \pmb{\checkmark} & \checkmark & \\ 
\hline

\multirow{2}{*}{Non-transformer Architecture~\S\ref{sec: efficient non-transformer}} 
& Modular Network & \pmb{\checkmark} & \checkmark & \checkmark & \checkmark & \\ 
\cline{2-7}
& Other Architecture & \pmb{\checkmark} & \checkmark & \checkmark & \checkmark & \\ 
\hline

\multirow{2}{*}{Distributed Training~\S\ref{sec: distributed training}} 
& Data Parallelism & \checkmark & \pmb{\checkmark} & \checkmark & & \\ 
\cline{2-7}
& Model Parallelism & \checkmark & \pmb{\checkmark} & \checkmark & & \\ 
\hline

Mixed-precision Training~\S\ref{sec: mixed precision training} & & \checkmark & \pmb{\checkmark} & \checkmark & & \checkmark \\ 
\hline

\multirow{2}{*}{Data Efficiency~\S\ref{sec: data efficiency}}
& Training Objective &  &  & \checkmark & \checkmark & \\ 
\cline{2-7}
& Data Augmentation & & & \checkmark & \checkmark & \\ 
\hline

\multirow{2}{*}{\shortstack{Parameter-efficient \\ Fine-tuning~\S\ref{sec: parameter efficient finetuning}}} 
& Adapter-based Fine-tuning & \checkmark & \pmb{\checkmark} & & & \\ 
\cline{2-7}
& Masking-based Fine-tuning & \checkmark & \pmb{\checkmark} & & & \\ 
\hline

Full-parameter Fine-tuning~\S\ref{sec: full parameter finetuning} & & \checkmark & \pmb{\checkmark} & \checkmark & & \\ 
\hline

\multirow{3}{*}{Pruning~\S\ref{sec: pruning}} 
& Unstructured Pruning & \pmb{\checkmark} & \pmb{\checkmark} & \checkmark & & \\ 
\cline{2-7}
& Structured Pruning & \pmb{\checkmark} & \pmb{\checkmark} & \pmb{\checkmark} & & \\ 
\cline{2-7}
& Contextual Pruning & \pmb{\checkmark} &   & \pmb{\checkmark} & & \\ 
\hline

\multirow{3}{*}{Quantization~\S\ref{sec: quantization}} 
& Weight Quantization & \checkmark & \pmb{\checkmark} & \pmb{\checkmark} & & \checkmark \\ 
\cline{2-7}
& Activation Quantization & \checkmark & \pmb{\checkmark} & \pmb{\checkmark} & &  \\ 
\cline{2-7}
& Fixed-point Quantization & \checkmark & \pmb{\checkmark} & \pmb{\checkmark} & & \\ 
\hline

\multirow{2}{*}{Knowledge Distillation~\S\ref{sec: knowledge distillation}} 
& White-box distillation & \checkmark & \pmb{\checkmark} & & & \\ 
\cline{2-7}
& Black-box distillation & \checkmark & \pmb{\checkmark} & & & \\ 
\hline

Low-rank Approximation~\S\ref{sec: low rank approximation} & & \pmb{\checkmark} & \pmb{\checkmark} & & & \\ 
\hline

\multirow{3}{*}{Dynamic Inference~\S\ref{sec: dynamic inference}} 
& Early Exit & \pmb{\checkmark} & & \checkmark & \checkmark & \\ 
\cline{2-7}
& Input Pruning & \pmb{\checkmark} & & & \checkmark & \\ 
\cline{2-7}
& Token Parallelism & \pmb{\checkmark} & & & \checkmark & \\ 
\hline

\multirow{2}{*}{Deployment Optimization~\S\ref{sec: deployment optimization}} 
& Hardware Offloading &  & \pmb{\checkmark} & & \pmb{\checkmark} & \\ 
\cline{2-7}
& Collaborative Inference & \checkmark & \pmb{\checkmark} & \checkmark & \pmb{\checkmark} &  \\ 
\hline

\end{tabular}
\caption{Mapping of Resource Efficiency Techniques to Key Resources in Large Language Models: This table presents a detailed overview of various techniques employed in the optimization of LLMs, categorizing them by their main and sub-categories. Bold checkmarks (\pmb{\checkmark}) indicate direct impacts and regular checkmarks (\checkmark) indicate indirect impacts on the resources.}
\vspace{-3mm}
\label{tab: technique resource mapping}
\end{sidewaystable}

In this section, we explore various techniques applied to Large Language Models (LLMs) to enhance their efficiency in using different resources. The focus is on five key resources: computation, memory, energy, financial cost, and network communication. Each technique discussed plays a vital role in optimizing resource efficiency for LLMs. The extent of their impact, whether direct or indirect, varies based on the resource in question. The table \ref{tab: technique resource mapping} provides a comprehensive mapping of these relationships.

\subsection{Computation efficiency}
\label{subsec:computation-efficiency}

Computation efficiency in LLMs is crucial for faster training and inference. Techniques like transformer architectures with approximated and hardware-aware attention directly enhance computation efficiency by reducing the complexity of operations. Approximated attention mechanisms, for instance, simplify the computationally intensive attention calculations, thus speeding up the process. Hardware-aware optimizations tailor models to exploit specific hardware capabilities, leading to more efficient computation. Unstructured, structured, and contextual pruning also directly impacts computation efficiency by eliminating redundant computations through the removal of less important weights or neurons. Indirect impacts are observed in data parallelism and parameter-efficient fine-tuning, where the distributed workload and reduced parameter updates respectively contribute to overall computational efficiency, albeit as a secondary effect.

\subsection{Memory efficiency}
\label{subsec:memory-efficiency}

Memory efficiency is critical, particularly for deployment on resource-constrained devices. Techniques like pruning and quantization explicitly target memory efficiency by reducing the model size. Pruning methods eliminate unnecessary weights, and quantization reduces the precision of weights, both leading to substantial memory savings. Knowledge Distillation, where a smaller model is trained to mimic a larger one, also directly improves memory efficiency. Indirect contributions come from distributed training, where data and model parallelism effectively manage memory usage across multiple devices, reducing the burden on individual units.

\subsection{Energy efficiency}
\label{subsec:energy-efficiency}

Energy efficiency is increasingly important in the context of sustainable AI. Structured pruning and quantization are directly beneficial, as they reduce the number of operations and the data size, leading to lower energy consumption for both training and inference. Contextual pruning, which adapts the pruning process based on the context, also leads to energy savings by minimizing unnecessary computations. While primarily aimed at computational efficiency, techniques like approximated attention indirectly contribute to energy savings due to the reduced computational load.

\subsection{Financial cost efficiency}
\label{subsec:financial-cost-efficiency}

Financial cost efficiency is an indirect but significant benefit of various resource-efficient techniques. Data efficiency methods, such as optimized training objectives and data augmentation, indirectly reduce costs by improving the effectiveness of the data used, leading to potentially shorter training times and less computational resource usage. Dynamic inference techniques like early exit and input pruning indirectly contribute to monetary efficiency by reducing the operational demands during the inference phase, which can lower the overall deployment costs. In addition to those techniques introduced earlier, some recent works exclusively minimize the monetary cost. For example, SpotServe~\cite{miao2023spotserve} is a novel system for serving generative large language models (LLMs) on preemptible GPU instances in the cloud, which are more cost-effective but less stable than regular instances. SpotServe dynamically adjusts LLM parallelization to manage instance availability and workload fluctuations, optimizing for throughput, latency, and cost. It also employs a unique approach for instance migration, minimizing communication costs, and utilizes stateful inference recovery to efficiently resume operations after preemptions, ultimately reducing costs by 54\% and improving latency compared to existing systems.

\subsection{Network communication efficiency}
\label{subsec:network-communication-efficiency}

In distributed training environments, network communication efficiency becomes crucial. Mixed-precision training explicitly addresses this by reducing the size of data that needs to be communicated between processors, directly impacting the efficiency of data transfer. Techniques like weight quantization also have a direct impact by minimizing the data payload during communication. Collaborative inference, an indirect approach, enhances communication efficiency by distributing inference tasks in a manner that optimizes data transfer and processing across the network.

\section{Benchmark and evaluation metrics}
\label{sec: dataset and metrics}

\subsection{Evaluation metrics}
Evaluating the resource efficiency of large language models (LLMs) involves considering a multifaceted range of metrics. We provide a comprehensive analysis of various metrics in this section. These metrics collectively offer a holistic understanding of the resource efficiency of large language models and are crucial for guiding model selection based on specific application requirements.

\subsubsection{Computation}
\begin{itemize}[leftmargin=*]
    \item \textbf{FLOPs} (Floating-point operations) represents the number of arithmetic operations on floating-point numbers, providing a quantifiable measure of the computation efficiency. In the context of LLMs, FLOPs capture the complexity of their internal computations, including attention mechanisms, feed-forward layers, and activation functions ~\cite{huang2023towards,jiao2019tinybert}.
    \item \textbf{Training time} refers to the total duration required to train an LLM, typically measured in wall-clock minutes, hours, or days ~\cite{dao2022flashattention,katharopoulos2020linearTrans}. It reflects the model's complexity and reveals the efficiency of the training algorithms and hardware. Optimized algorithms and hardware can significantly reduce training time, making LLMs more accessible and affordable.
    \item \textbf{Inference time/latency} quantifies the time it takes for an LLM to generate an output after receiving an input. It is typically measured in wall-clock time or CPU/GPU/TPU clock time in milliseconds or seconds and can be accessed in two ways: 1) end-to-end latency ~\cite{kwon2023vLLM} and 2) next token generation latency ~\cite{shen2023efficient}. The inference time is crucial for evaluating the practical applicability of LLMs in real-world scenarios, particularly those with time-sensitive interactions. 
    \item \textbf{Throughput} quantifies the model's efficiency in processing requests. It defines the rate at which LLMs can generate output tokens or complete tasks, typically measured in tokens per second (TPS) ~\cite{borzunov2022petals,kwon2023vLLM} or queries per second ~\cite{wu2022efficient, ma2021dynaboard}. A high throughput indicates an LLM's ability to handle multiple requests quickly, making it suitable for high-volume applications like real-time chatbots or large-scale content generation.
    \item \textbf{Speed-up ratio} measures the improvement in inference speed compared to a baseline model, typically expressed as a multiple (e.g., 1.2x, 3.5x) ~\cite{wang2021lightseq,FasterTransformer}. The ratio is calculated as the quotient of the baseline execution time to the time taken by the improved LLM or the quotient of the improved LLM’s throughput to that of the baseline model. The speed-up ratio draws a relative comparison between the two models, which offers a quantitative measure of progress in optimizing model architectures, training algorithms, and hardware optimizations.

\end{itemize}

\subsubsection{Memory}
\begin{itemize}[leftmargin=*]
    \item \textbf{Number of parameters} represents the number of adjustable variables in the LLM's neural network. A higher number of model parameters generally indicates a more complex model with a greater capacity to learn and represent intricate patterns in the data~\cite{vaswani2017attention,artetxe2022efficient}. However, this complexity often comes at a cost of increased computational and memory requirements.
    \item \textbf{Model size} of an LLM is determined by the storage space required for storing the entire model. It directly correlates with the number of parameters and is also influenced by the model’s architecture, training data size, and any additional resources it needs to run. The model size is often measured by the peak memory usage during training or inference ~\cite{zandieh2023kdeformer,rabe2021memEffAttn}, which can be easily obtained through system monitoring tools. Similar to the number of parameters, larger architectures, and more data have been able to continuously improve transformer models’ performances ~\cite{devlin2018bert}, but this can become a limiting factor, especially when dealing with extremely large models that may not fit into the memory of standard hardware.
\end{itemize}

\subsubsection{Energy}
\begin{itemize}[leftmargin=*]
    \item \textbf{Energy consumption} of an LLM is typically expressed in Watt-hours (Wh) or Joules (J), reflecting the electrical power used during the LLM's lifecycle. ~\cite{strubell2019energy} proposed a method to calculate the total power consumption by combining GPU, CPU, and DRAM consumption and multiplying it by the Power Usage Effectiveness (PUE), accounting for the additional energy required to support the entire compute infrastructure.
    \item \textbf{Carbon emission} captures the greenhouse gas emissions associated with the model's energy usage. It can be calculated by multiplying the energy consumption (kWh) by the carbon intensity, the grams of carbon dioxide or equivalents emitted per kWh of energy used ($gCO_{2eq}/ kWh$) by the local power grid. 
\end{itemize}

CodeCarbon ~\cite{schmidt2021codecarbon}, carbontracker ~\cite{anthony2020carbontracker}, and experiment-impact-tracker ~\cite{henderson2020towards} are available software packages designed for real-time tracking of energy consumption and carbon emissions. Other tools like MLCO2 Impact \cite{lacoste2019quantifying} and LLMCarbon ~\cite{faiz2023llmcarbon} leverage machine learning to predict the energy usage and carbon footprint before actual training, enabling more informed resource allocation. 

However, focusing solely on training overlooks the full picture. It is encouraged to report the energy consumption and carbon footprint during all stages of an LLM’s lifecycle: while energy consumption during the training and development phases can be substantial, its significance may diminish when compared to the cumulative lifetime costs of inference if the model is utilized intensively in production. BLOOM ~\cite{scao2022bloom} highlights this holistic perspective, demonstrating the significant impact of life cycle assessment in accurately quantifying environmental costs.

\subsubsection{Financial cost}
\begin{flushleft}
While the financial cost of developing and deploying LLMs is rarely reported by researchers, understanding it is valuable for prioritizing research directions based on cost-effectiveness, encouraging collaboration to optimize resource allocation, and promoting responsible LLM development. We propose a novel metric \textit{dollars per parameter} for this measure.
\end{flushleft}

\begin{itemize}[leftmargin=*]
    \item  \textbf{Dollars per parameter} is a scaled metric for the financial cost of an LLM. It is obtained by dividing the total cost of training (or running) the LLM by the number of parameters. It accounts for differences in model size and complexity, allowing for fairer comparisons of cost efficiency between different models. Researchers may consider reporting not only the cloud compute and electricity cost ~\cite{strubell2019energy} and the hardware and software expenses associated with training and running, but also the personnel costs involved in data collection, model architecture, and fine-tuning.
\end{itemize}

\subsubsection{Network communication}
\begin{itemize}[leftmargin=*]
    \item \textbf{Communication volume} refers to the total amount of data (in megabytes, gigabytes, etc.) transmitted across the network during a specific LLM execution or training run ~\cite{wu2023rethinking}. Measuring it involves monitoring network traffic between connected nodes, often employing dedicated software or system logs. Monitoring the communication volume is important for distributed LLMs, as high communication costs can lead to bottlenecks, slowing down training and increasing infrastructure expenses.
\end{itemize}

\subsubsection{Other metrics}
\begin{itemize}[leftmargin=*]
    \item \textbf{Compression ratio} quantifies the reduction in size of the compressed model compared to the original model. It can be expressed in the percentage of size reduced ~\cite{han2015deep,tao2023structured} or the percentage of weights remaining ~\cite{li2023losparse}. High compression ratios without significantly compromising the performance indicate better compression efficiency, which facilitates deployment across diverse platforms, including those with limited computational resources.
    \item \textbf{Loyalty} and \textbf{Fidelity} are two similar metrics proposed by ~\cite{xu2021beyond} and ~\cite{stanton2021does} respectively to measure the resemblance between the teacher and student models in terms of both predictions consistency and predicted probability distributions alignment. For large teacher models, enhancements in fidelity lead to improvements in generalization: the student's ability to predict previously unseen, in-distribution data ~\cite{stanton2021does}. Achieving a high loyalty (or fidelity) is also important for preserving a model’s fairness, as recent work showed that model compression can amplify existing algorithmic bias ~\cite{hooker2020characterising}.
    \item \textbf{Robustness} is another important but rarely reported metric. ~\cite{xu2021beyond} used two metrics: after-attack accuracy and query number to evaluate robustness. After-attack accuracy measures the model's post-attack performance, while the query number reflects the complexity of the attack required to succeed. Evaluating the robustness of LLMs is crucial to ensure reliability in real-world applications, especially for compressed models, as ~\cite{su2018robustness} found smaller deep neural networks tend to be more vulnerable to adversarial attacks, where slight input modifications can manipulate their output.

    \item \textbf{Pareto optimality} ~\cite{treviso2023efficient} is achieved by striking an optimal balance between various competing factors. It offers a systematic approach to balance trade-offs between resource efficiency and task performance by identifying solutions that reside on the Pareto frontier ~\cite{pareto1964cours} — the boundary that delineates optimal trade-offs. This concept is especially meaningful as it provides a principled framework for decision-making, allowing researchers to navigate the intricate landscape of LLMs and make informed choices. For example, ~\cite{santhanam2022moving} examined the trade-offs between cost and accuracy with the Pareto frontier, identifying the most practical real-world information retrieval models. \cite{liu2021towards} explored the Pareto frontier between performance and FLOPs to determine whether and how much a method achieves Pareto improvement.
    
\end{itemize}

\subsection{Benchmarks}
The evaluations of LLMs’ resource efficiency currently rely heavily on general NLP benchmarks\footnote{For an extensive collection of general NLP benchmarks, refer to a comprehensive overview by ~\cite{naveed2023comprehensive}.}, such as GLUE ~\cite{wang2018glue}, SuperGLUE ~\cite{wang2019superglue}, WMT ~\cite{bojar2016findings,barrault2020findings}, and SQuAD ~\cite{rajpurkar2016squad,rajpurkar2018know}. While the existing general NLP benchmarks offer valuable insights into a model’s performance on various tasks, they often fail to capture the nuances of resource-efficient approaches. We present a few benchmarks that involve efficiency considerations; however, there is still a need for more comprehensive and specialized benchmarks for LLM efficiency evaluations. These benchmarks should go beyond traditional metrics and address the unique challenges associated with the efficient utilization of computational resources, memory, energy consumption, financial cost, and communication overhead.

\begin{itemize}[leftmargin=*]
    \item \textbf{Dynaboard} ~\cite{ma2021dynaboard} is an open-source dynamic benchmark that allows users to submit NLP models to be evaluated in the cloud, enabling real-time interaction and a more comprehensive assessment of model quality. Submitted models are evaluated on a combination of datasets across four tasks: Natural Language Inference, Question Answering, Sentiment Analysis, and Hate Speech. In addition to accuracy, Dynaboard also collects additional metrics such as memory usage, throughput, fairness, and robustness. Models are ranked according to a novel utility-based aggregation of these metrics called Dynascore, which can be customized by leaderboard creators by assigning weights that reflect the relative importance of each metric. 
    \item \textbf{EfficientQA} ~\cite{min2021neurips} is an open-domain Question Answering (QA) challenge at NeurIPS 2020\footnote{https://efficientqa.github.io/} that focuses on building accurate, memory-efficient QA systems. It promotes efficient memory usage through three restrained tracks based on model size and accuracy: the most accurate model under 6 GB, the most accurate model under 500 MB, and the smallest model that achieves 25\% accuracy. The model size is measured as the Docker image size that contains the complete, self-contained question-answering system. 
    \item \textbf{SustaiNLP 2020\footnote{https://sites.google.com/view/sustainlp2020} Shared Task} ~\cite{wang2020overview} encourages participants to develop energy-efficient NLP models. It uses SuperGLUE ~\cite{wang2019superglue} to access the model's performance across eight diverse NLU tasks. In addition to standard SuperGLUE metrics, this challenge uniquely evaluates the energy consumption of each submission during inference using ~\cite{henderson2020towards}'s experiment-impact-tracker.
    \item \textbf{ELUE} (Efficient Language Understanding Evaluation) ~\cite{liu2021towards} is a benchmark and platform designed to evaluate and compare the efficiency of various NLP models. It covers six NLP datasets spanning Sentiment Analysis, Natural Language Inference, Similarity, and Paraphrase tasks. ELUE supports online evaluation for model performance, average FLOPs, and the number of parameters. While Dynaboard ~\cite{ma2021dynaboard} and EfficientQA \cite{min2021neurips} submissions require the containerized model along with its required environment, ELUE only requires the model definition file in Python, which is less costly for users to upload. 
    
    \item \textbf{VLUE} (Vision-Language Understanding Evaluation) ~\cite{zhou2022vlue} is a multi-task multi-dimension benchmark for evaluating vision-language model (VLM)s. It covers a set of fundamental vision language tasks: Image-Text Retrieval, Visual Question Answering, Visual Reasoning, and Visual Grounding, and maintains an online platform and leaderboard for evaluation and comparison of vision-language pre-training (VLP) models. Notably, VLUE measures both performance and inference time and evaluates the efficiency-performance trade-off, providing a more comprehensive assessment of VLMs' practical value.
    \item \textbf{Long-Range Arena} (LRA)\footnote{https://github.com/google-research/long-range-arena} ~\cite{tay2020long} is a benchmark suite specifically designed for evaluating the performance of efficient Transformer models on long-context tasks. It features a variety of tasks that require reasoning over long contexts, ranging from 1,000 to 16,000 tokens in length. These tasks cover different modalities like text, natural language, synthetic images, and mathematical expressions, and involve various reasoning types like similarity, structural, and visual-spatial reasoning. Evaluations can be run under controlled resource constraints like memory budget or execution time limit. This forces models to be efficient within these limitations, highlighting their ability to optimize performance under real-world constraints.
    \item \textbf{Efficiency-aware MS MARCO} is a post-hoc leaderboard created by ~\cite{santhanam2022moving} for the MS MARCO information retrieval (IR) benchmark ~\cite{bajaj2016ms} to include efficiency metrics such as average per-query latency and the corresponding cost budget in addition to accuracy to provide a more holistic evaluation of IR systems. Following ~\cite{ma2021dynaboard}, ~\cite{santhanam2022moving} ranked the models using an aggregation of accuracy, latency, and cost, expressed as a Dynascore. 
    %\item \textbf{Unseen Instructions Datasets}
\end{itemize}

\section{Open challenges and future directions}
\label{sec: challenges}

As the field of Large Language Models (LLMs) continues to advance, it faces a multitude of open challenges that pave the way for promising research directions. This section elaborates on these challenges and potential avenues for future exploration.

\subsection{Managing resource type disagreements}
A complex challenge in optimizing Large Language Models (LLMs) is the reconciliation of conflicting resource demands. Various optimization techniques introduce trade-offs between different performance metrics~\cite{zhang2023pruning,hsieh2023distilling,frantar2022gptq,liu2023deja}. For instance, some methods might enhance computational efficiency at the cost of increasing the number of model parameters or vice versa~\cite{hu2021lora,dettmers2023qlora}. Additionally, strategies that reduce computational operations could inadvertently lead to higher memory requirements due to the complexities involved in managing sparse data structures~\cite{frantar2023massive,liu2023deja}. The key challenge, therefore, is to develop a holistic optimization strategy for LLMs. This strategy should balance multiple objectives, including computational efficiency, parameter count, and memory usage, ensuring that improvements in one area do not disproportionately compromise performance in another. Achieving this balance is essential for the advancement of LLMs, making them both efficient and practical for diverse applications.

\subsection{Combining techniques for resource efficiency}
In the realm of Large Language Models (LLMs), a significant challenge lies in effectively amalgamating the diverse array of available methods to enhance overall resource efficiency. While numerous techniques exist to optimize different aspects of LLMs, there's a notable scarcity of research on how these methods can be cohesively combined. For instance, Ford et al.~\cite{ford2021portauthority} propose a methodology to integrate energy efficiency analysis into software development cycles, which is relevant for the development of efficient LLMs. Similarly, Xie et al.~\cite{xie2019integration} discuss the integration of resource allocation and task assignment for optimizing business processes, a concept that can be applied to the management of LLM resources. An integrative approach that systematically brings together various strategies could markedly improve the efficiency of these large models. This approach could draw from the insights of~\cite{shamshirband2021game}, who surveyed computational intelligence-based optimization and game theory approaches for resource allocation in computing environments. This could inform the development of strategies for efficient resource utilization in LLMs.

\subsection{Standardized and unified evaluation}
A critical challenge in the realm of LLMs is the absence of universally accepted benchmarks specifically tailored for evaluating the resource efficiency of these models. While several benchmarks exist for assessing aspects like model compression and acceleration~\cite{ma2021dynaboard,min2021neurips}, they fall short of providing a comprehensive and consistent evaluation of resource efficiency. This shortcoming is due to varying factors such as differences in speed-up ratios, the number of parameters, accuracy levels, and even the type of hardware used in different studies. These discrepancies highlight the urgent need for standardized benchmarks that are designed with a focus on resource efficiency. Such benchmarks should facilitate direct comparisons and enable more accurate and holistic analyses of how various LLMs utilize computational resources, including aspects like energy consumption, memory usage, and processing power. Establishing these benchmarks is essential for advancing the development of more resource-efficient LLMs, a key priority given the increasing size and complexity of these models.

\subsection{Explainability and robustness}
The pursuit of efficiency in Large Language Models (LLMs) brings to the fore concerns about their explainability and robustness, echoing issues identified in earlier research on pre-trained language models. For instance, while some techniques significantly improve performance in reasoning tasks, there is often a lack of clarity regarding the underlying reasons for their effectiveness~\cite{wei2022chain,diao2023active}. This obscurity highlights the need for integrating explainable approaches into the development of efficient LLMs. Such integration would address the dual needs of interpretability and simplified evaluation, thus enhancing the reliability and predictability of LLMs in real-world applications. The goal is to develop methods that not only optimize resource use but also maintain a level of transparency and resilience, ensuring that these models can be trusted and easily understood in various deployment scenarios.

\subsection{AutoML for resource-efficient LLMs}
The integration of Automated Machine Learning (AutoML) into the development of resource-efficient Large Language Models (LLMs) represents a burgeoning field of interest. Traditional methods for enhancing resource efficiency in LLMs, such as knowledge distillation, pruning, weight sharing, and low-rank factorization, typically rely on expert-driven heuristics and intricate manual interventions~\cite{zhang2023pruning,liu2023deja,yang2023quantization}. For instance, designing effective loss functions for knowledge distillation or determining saliency scores for pruning involves a considerable amount of human judgment and expertise~\cite{frantar2023massive,hsieh2023distilling}. To mitigate this reliance on human input, there's a growing emphasis on applying techniques like Meta-Learning~\cite{vilalta2002perspective} and Neural Architecture Search (NAS)~\cite{elsken2019neural}. These AutoML strategies show promise in automating aspects of model optimization. By doing so, they could significantly reduce the need for manual hyperparameter tuning and bespoke model design, potentially leading to more efficient and easily customizable LLMs. This approach not only streamlines the optimization process but also opens avenues for more innovative and adaptable efficiency solutions in the realm of LLMs.

\subsection{Edge computing with LLMs}
Deploying Large Language Models (LLMs) in edge computing environments presents unique challenges due to the inherent limitations of edge devices. These devices often face constraints in terms of battery life, computational power, and memory resources~\cite{murshed2021machine,shi2016edge}. Additionally, issues such as data privacy and network latency further complicate their use~\cite{zhang2021survey}. To address these challenges, there is a need to develop LLM techniques that are not only resource-efficient but also mindful of privacy concerns. Key to this development is the ability to facilitate effective on-device training and operational capabilities of LLMs, making them viable for a range of practical applications in edge computing scenarios.

\subsection{Theoretical insights into scaling laws}
A crucial yet underexplored area in the realm of Large Language Models is the theoretical understanding of their scaling laws and generalization capabilities. A deeper insight into how LLMs' performance scales with their size and complexity is essential. Such understanding is pivotal in developing methods that are not just focused on model compression but are tailored to enhance the overall resource efficiency of LLMs. Gaining this knowledge is fundamental for researchers to more effectively explore and innovate within the design space of LLMs, ultimately leading to solutions that are both powerful and resource-efficient. This theoretical foundation is key to advancing the field and optimizing LLMs for a wider range of applications and environments.

\section{Conclusion}
\label{sec: conclusion}

In this survey, we have systematically explored the realm of resource-efficient Large Language Models (LLMs), offering a comprehensive view of the current state-of-the-art techniques and methodologies. We started by providing a foundational understanding of the challenges and necessities in developing resource-efficient LLMs in Section~\ref{sec: preliminary and taxonomy}. This set the stage for a deeper investigation into various aspects of LLMs, including architecture design, pre-training, fine-tuning, inference, and system design.

In Section~\ref{sec: architecture design}, we highlighted innovative approaches in LLM architecture that contribute to resource efficiency, emphasizing the significance of both transformer and non-transformer architectures. Sections~\ref{sec: pretrain}, \ref{sec: finetune}, and \ref{sec: inference} delved into the nuances of pre-training, fine-tuning, and inference phases of LLMs, respectively, showcasing how each stage presents unique opportunities and challenges for resource optimization.

Our exploration in Section~\ref{sec: system design} underscored the importance of holistic system design in achieving resource efficiency, where both hardware and software aspects play a crucial role. In Section~\ref{sec: application}, we examined the practical applications and evaluations of these techniques, linking them back to the resource taxonomy established earlier in the survey.

The survey also shed light on the current benchmarks and evaluation metrics in Section~\ref{sec: dataset and metrics}, which are crucial for quantifying and comparing the efficiency of various LLMs. In Section~\ref{sec: challenges}, we identified key open challenges and future directions, pointing towards unexplored areas and potential breakthroughs that could further advance the field of resource-efficient LLMs.

As we conclude, it is evident that while significant progress has been made in developing resource-efficient LLMs, there remains a vast landscape of opportunities for further research and innovation. The insights and discussions presented in this survey aim to serve as a guiding framework for future work in this rapidly evolving field, with the ultimate goal of achieving highly efficient LLMs that are accessible and sustainable for a wide range of applications.

\bibliography{references}% common bib file

\end{document}